\documentclass[letterpaper, 10 pt, conference]{ieeeconf}  
\IEEEoverridecommandlockouts  
\usepackage{graphicx}
\usepackage{mathptmx}      
\usepackage{array}
\usepackage[colorlinks,bookmarksopen,bookmarksnumbered,citecolor=blue,urlcolor=blue]{hyperref}
\usepackage{enumerate}
\usepackage{amsfonts}
\usepackage{subcaption}
\usepackage{caption}
\usepackage{mathtools, cuted}
\newcolumntype{L}[1]{>{\raggedright\let\newline\\\arraybackslash\hspace{0pt}}m{#1}}
\newcolumntype{C}[1]{>{\centering\let\newline\\\arraybackslash\hspace{0pt}}m{#1}}
\newcolumntype{R}[1]{>{\raggedleft\let\newline\\\arraybackslash\hspace{0pt}}m{#1}}

\newcommand{\ie}{\textit{i.e.}}
\newcommand{\eg}{\textit{e.g.}}

\title{Robot-to-Robot Relative Pose Estimation using Humans as Markers}

\author{Md Jahidul Islam$^{1}$, Jiawei Mo$^{2}$ and Junaed Sattar$^{3}$%
	\thanks{The authors are with the Interactive Robotics and Vision Laboratory, 
		Department of Computer Science and Engineering, Minnesota Robotics Institute (MnRI), 
		University of Minnesota, Twin Cities, US \newline
		{\tt\small E-mail:$\{ ^{1}$islam034,  $^{2}$moxxx066, $^{3}$junaed$\}$@umn.edu} }
}

\begin{document}

\maketitle
\thispagestyle{empty}
\pagestyle{empty}

\begin{abstract}
\textit{In this paper, we propose a method to determine the 3D relative pose of pairs of communicating robots by using human pose-based key-points as correspondences. We adopt a `leader-follower' framework, where at first, the leader robot visually detects and triangulates the key-points using the state-of-the-art pose detector named OpenPose. Afterward, the follower robots match the corresponding 2D projections on their respective calibrated cameras and find their relative poses by solving the perspective-n-point (PnP) problem. In the proposed method, we design an efficient person re-identification technique for associating the mutually visible humans in the scene. Additionally, we present an iterative optimization algorithm to refine the associated key-points based on their local structural properties in the image space. We demonstrate that these refinement processes are essential to establish accurate key-point correspondences across viewpoints. Furthermore, we evaluate the performance of the proposed relative pose estimation system through several experiments conducted in terrestrial and underwater environments. Finally, we discuss the relevant operational challenges of this approach and analyze its feasibility for multi-robot cooperative systems in human-dominated social settings and feature-deprived environments such as underwater.}
\end{abstract}

\section{Introduction}
Accurate computation of relative pose is essential in multi-robot estimation problems such as cooperative tracking, localization~\cite{rekleitis2002multi}, planning, mapping~\cite{se2005vision}, and more. Unless global positioning information (\eg, GPS, USBL) is available, the robots need to estimate their positions and orientations relative to each other based on their exteroceptive sensory measurements and noisy odometry~\cite{zhou2008robot}. This process is necessary for registering their measurements to a common frame of reference in order to maintain coordination during task execution.

\begin{figure}[ht]
\centering
\includegraphics [width=\linewidth]{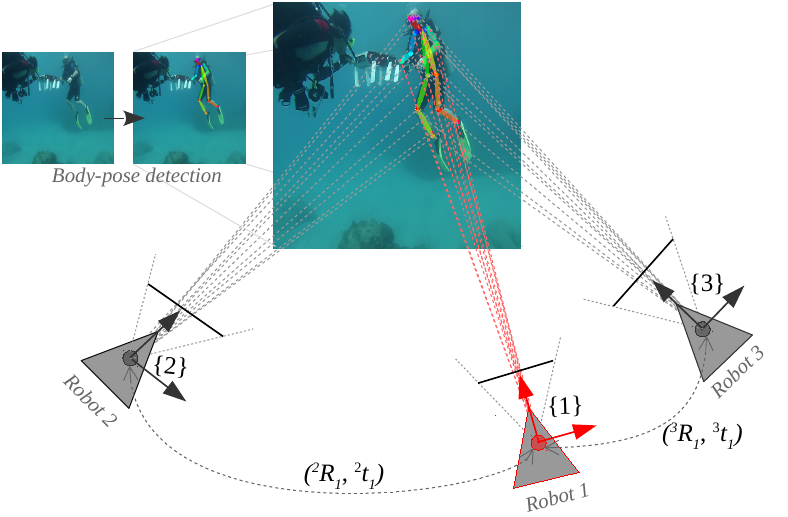} 
\caption{\footnotesize A simplified illustration of 3D relative pose estimation between robot 1 and robot 2 (3). The robots know the transformations between their intrinsically-calibrated cameras and respective global frames, \ie, \{1\}, \{2\}, and \{3\}. Robot 1 is considered as the leader (equipped with a stereo camera) and its pose in global coordinates ($^1R_G$, $^1t_G$) is known. Robot 2 (3) finds its unknown global pose by cooperatively localizing itself relative to robot 1 using the human pose-based key-points as common landmarks.}%
\label{fig:1}
\end{figure}%

In a cooperative setting, robots with visual sensing capabilities solve the relative pose estimation problem by triangulating mutually visible local features and landmarks. A lack of salient features significantly affects the accuracy of this estimation~\cite{valgren2010sift}, which eventually hampers the overall success of the operation. Such difficulties often arise in poor visibility conditions underwater due to a lower number of salient features and natural landmarks~\cite{damron2018underwater,sattar2008enabling}. Nevertheless, proximity of human divers to robots is a fairly common occurrence in applications~\cite{islam2018understanding} at shallow depths. Besides, humans are frequently present and visible in many social scenarios~\cite{islam2018person,kummerle2013navigation} where natural landmarks are not reliably identifiable due to repeated textures, noisy visual conditions, etc. Hence, the problem of having limited natural landmarks can be alleviated by using mutually visible humans as \emph{markers} (\ie, features correspondences), particularly in human-robot collaborative applications. Despite the potential, the feasibility of using human presence or body-pose for robot-to-robot relative pose estimation has not been explored in the literature.

In this paper, we propose a method for computing six degrees-of-freedom ($6$-DOF) robot-to-robot transformation between pairs of communicating robots by using mutually detected humans' pose-based key-points as correspondences. As illustrated in Fig.~\ref{fig:1}, we adopt a \emph{leader-follower} framework where one of the robots (equipped with a stereo camera) is assigned as a leader. First, the leader robot detects and triangulates 3D positions of the key-points in its own frame of reference. Then the follower robot matches the corresponding 2D projections on its intrinsically calibrated camera and localizes itself by solving the perspective-n-point (PnP) problem~\cite{zheng2013revisiting}. It is to be noted that this entire process of \emph{extrinsic calibration} is automatic and does not require prior knowledge about the robots' initial positions. Additionally, it is straightforward to extend the leader-follower framework for multi-robot teams from the pairwise solutions. Furthermore, if the leader robot has global positioning information, \ie, has a GPS or an USBL receiver, the follower robots can use that information to localize themselves in the global frame as well.

In addition to the conceptual design, we present an end-to-end system with efficient solutions to the practicalities involved in the proposed robot-to-robot pose estimation method. As mentioned, we use OpenPose~\cite{cao2017realtime} for detecting human body-poses in the image space. Although it provides reliable detection performance, the extracted 2D key-points across different views do not necessarily associate as a correspondence. We propose a twofold solution to this: 
\begin{itemize}
    \item First, we design an efficient person re-identification module by evaluating the hierarchical similarities of the key-point regions in the image space. It takes advantage of the consistent human pose structures across viewpoints and evaluates their pair-wise similarities for fast body-pose association. We also demonstrate that the state-of-the-art (SOTA) appearance-based person re-identification models fail to provide acceptable performance under single-board real-time constraints.  
    \item Subsequently, we formulate an iterative optimization algorithm to refine the noisy key-point correspondences by further exploiting their local structural properties in respective images. We demonstrate that the pair-wise key-point refinement is crucial to ensure their validity in a perspective geometric sense.     
\end{itemize}
This two-stage process facilitates efficient and robust key-point associations across viewpoints for accurate robot-to-robot relative pose estimation. In this paper, we primarily focus on these two novel modules because the rest of the computational aspects are generic to all multi-robot cooperative pose estimation systems. Nevertheless, we present a fast implementation of the proposed system and evaluate its end-to-end performance over several terrestrial and underwater field experiments. Lastly, we analyze its practical feasibility and discuss various operational considerations. 
\section{Related Work}

\subsection{Robot-to-robot Relative Pose Estimation}
The problem of robot-to-robot relative pose estimation has been thoroughly studied for 2D planar robots, particularly for range and bearing sensors. Analytic solutions for determining $3$-DOF robot-to-robot transformation using mutual distance and/or bearing measurements involve solving an over-determined system of nonlinear equations~\cite{zhou2008robot,trawny2010global}. Similar solutions for the 3D case, \ie, for determining 6-DOF transformation using inter-robot distance and/or bearing measurements, has been proposed as well~\cite{zhou2011determining,trawny2010interrobot}. In practice, these analytic solutions are used as an initial estimate for the relative pose, and then iteratively refined by optimization techniques (\eg~nonlinear weighted least-squares) to account for the noisy observation and uncertainty in robot motion.

Robots that rely on visual perception (\ie, use cameras as exteroceptive sensors) solve the relative pose estimation problem by triangulating mutually visible features and landmarks~\cite{wang19923d}. Therefore, it reduces to solving the PnP problem by using sets of 2D-3D correspondences between geometric features and their projections on respective image planes~\cite{zheng2013revisiting}. Although high-level geometric features (\eg, lines, conics) have been proposed, point-based features are typically used in practice for relative pose estimation~\cite{janabi2010kalman}. 
Moreover, the PnP problem is solved either using iterative approaches by formulating the over-constrained system ($n$ $>3$) as a nonlinear least-squares problem, or by using sets of three non-collinear points ($n=$ $3$) in combination with Random Sample Consensus (RANSAC) to remove outliers~\cite{fischler1981random}. Besides, vision-based approaches often use temporal-filtering methods, the extended Kalman-filter (EKF) in particular, to reduce the effect of noisy measurements in order to provide near-optimal pose estimates~\cite{wang19923d,janabi2010kalman}. On the other hand, it is also common to simplify the relative pose estimation by attaching specially designed calibration-patterns on each robot~\cite{rekleitis2006simultaneous}. However, this requires that the robots operate at a sufficiently close range, and remain mutually visible.

\subsection{Human Body-Pose Detection}
Visual detection of 2D human pose has made significant progress over the last decade. The SOTA methodologies can be categorized into the top-down and bottom-up approaches. The top-down approaches~\cite{gkioxari2014using,pishchulin2012articulated} detect the humans in the image space first, and then perform localization and association of their body-parts. One major limitation of these approaches is that their run-times are proportional to the number of persons in the image. Additionally, the robustness of the pose estimation largely depends on the accuracy of their person detectors. In contrast, the bottom-up approaches~\cite{cao2017realtime,pishchulin2016deepcut} do not suffer from these two issues. However, they require solving a more computationally challenging inference problem of learning global contextual cues for simultaneous body-part detection and association.

The classical approaches typically use pictorial structures~\cite{ferrari2008progressive,andriluka2009pictorial} to model the appearance of human body-parts. A set of densely sampled shape descriptors are used for localizing the body-parts and then classifiers such as AdaBoost, SVMs, etc., are used for detection. Associating the detected body-parts is rather challenging; a mixture of tree-based models are typically used to learn separate pairwise relationships for different body-part configurations~\cite{johnson2011learning}. Graph-based connectivity models are then used to formulate the inference (association) as a graph-cut problem. These pairwise connectivity models can be further generalized~\cite{pishchulin2013poselet} to capture the anatomical relationships among multiple body-parts. Recently proposed approaches use Deep Neural Networks (DNNs) to learn the human pose detection from large training datasets to perform fast and accurate global inference. DeepPose~\cite{toshev2014deeppose}, for instance, formulates the problem as a regression problem and uses a cascade of DNNs to learn the inference in a holistic fashion. On the other hand, OpenPose~\cite{cao2017realtime} jointly learns to detect and associate using \emph{pose machines}~\cite{ramakrishna2014pose}. In contrast to DNNs, each module of a pose machine is trained locally; the sequential predictions of these modules are then refined to perform a hierarchical joint inference. Such hierarchical structures facilitate fast inference for multi-person pose estimation in addition to achieving SOTA performance. Due to these compelling reasons, we use OpenPose in this work.

\subsection{Human-aware Robot Control}
Human-awareness is important for autonomous mobile robots operating in social settings and human-robot collaborative applications. A large body of literature and systems exist~\cite{islam2018understanding,mead2017autonomous} which focus on the areas of understanding human motion, instructions, behaviors, etc. Additionally, tracking human pose relative to a robot is particularly common in applications such as person tracking or following~\cite{islam2018person,montemerlo2002conditional}, collaborative manipulation~\cite{mainprice2013human}, behavior imitation~\cite{lei2015whole}, etc. However, the feasibility of using humans' presence or their body-poses as markers for robot-to-robot relative pose estimation has not been explored in the literature.

\section{System Design and Methodology}
\begin{figure}[h]
\centering
\includegraphics [width=\linewidth]{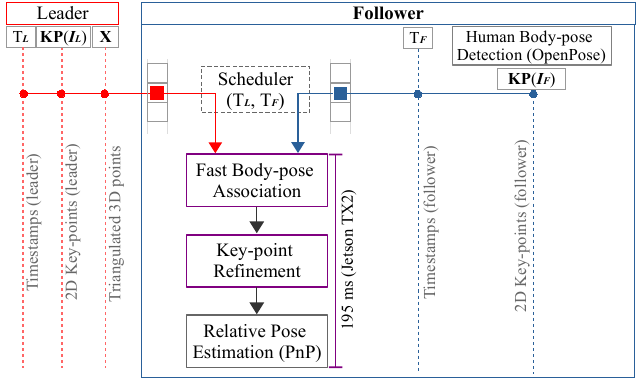}
\caption{\footnotesize The end-to-end computational pipeline is outlined from the perspective of a follower robot which shares a clock with the communicating leader robot by using a timestamp-based buffer scheduler for synchronized data registration. The mutually visible human body-pose based key-points are then associated and refined for relative pose estimation. We design these two novel components (marked in purple boxes) to establish robust and accurate key-point correspondences at a fast rate ($195$ milliseconds per estimation on a Nvidia Jetson TX2).
}%
\label{fig:sys}
\end{figure}

As shown in Fig.~\ref{fig:sys}, the proposed robot-to-robot relative pose estimation system incorporates several computational components: detection of human body-poses in images captured from different views (by leader and follower robots), pair-wise association of the detected humans across viewpoints, geometric refinement of the key-point correspondences, and 3D pose estimation of the follower robot relative to the leader. We present their methodological details and relevant design choices in the following sections.

\subsection{Human Body-Pose Detection}
OpenPose~\cite{cao2017realtime} is an open-source library for real-time multi-human 2D pose detection in images, originally developed using Caffe and OpenCV libraries\footnote{\url{github.com/CMU-Perceptual-Computing-Lab/openpose}}. We use a Tensorflow implementation\footnote{\url{github.com/ildoonet/tf-pose-estimation}} based on the \emph{MobileNet model} that provides faster inference compared to the original model (also known as the \emph{CMU model}). Specifically, it processes a $368\times368$ image in $180$ milliseconds on the embedded computing board named Jetson TX2~\cite{Jetson}, whereas the original model takes multiple seconds.

\begin{figure}
	\centering
	\includegraphics [width=0.595\linewidth]{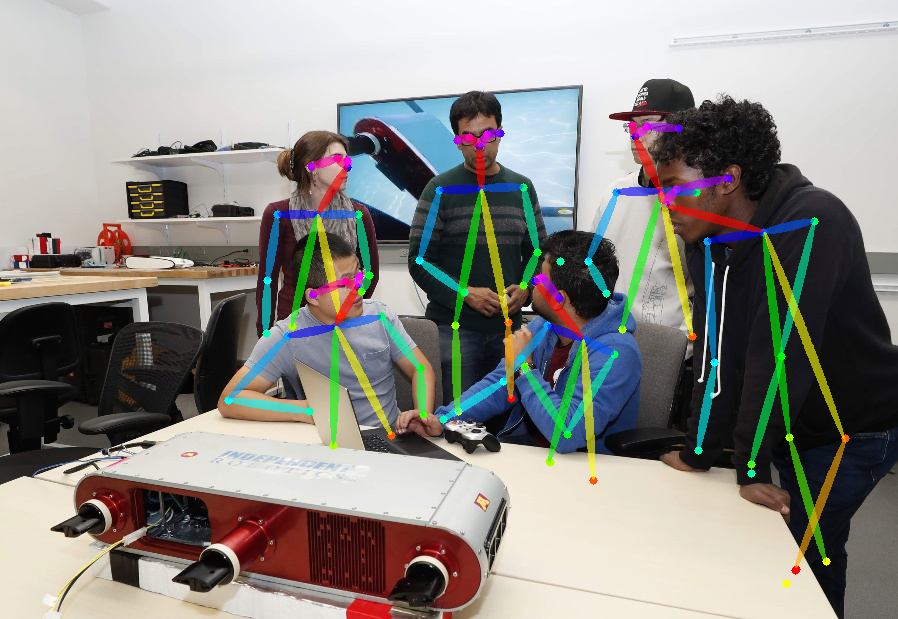} 
	\includegraphics [width=0.385\linewidth]{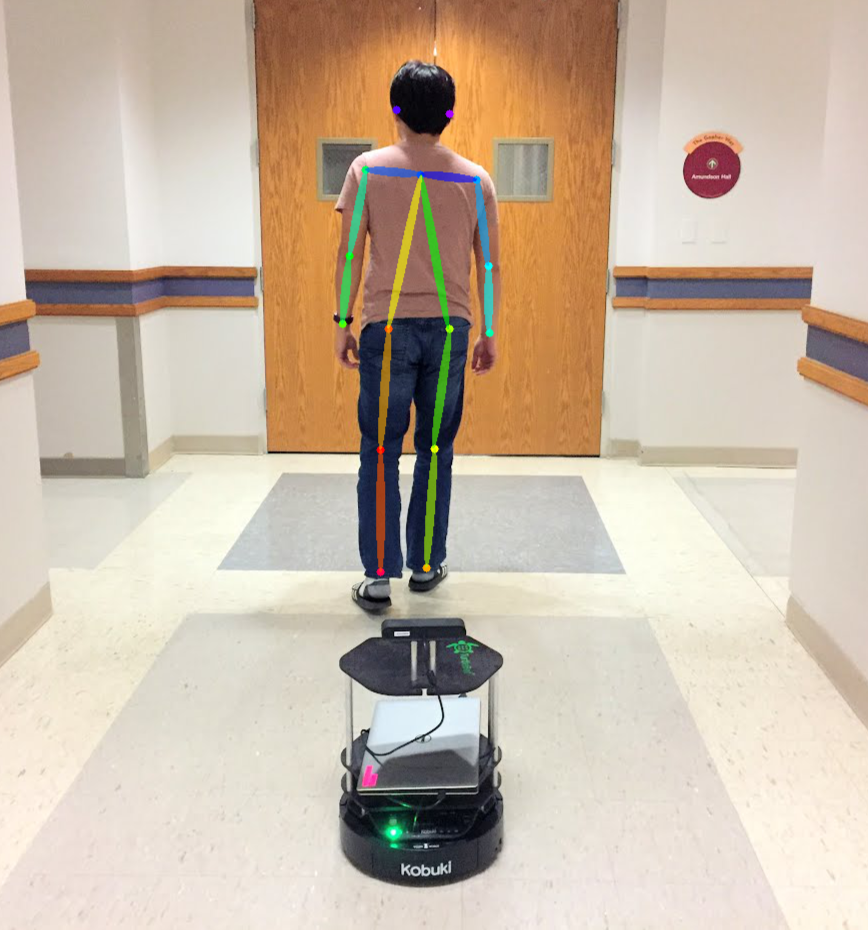} \\
	\vspace{1mm}
	\includegraphics [width=0.325\linewidth]{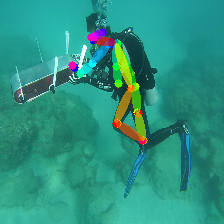} 
	\includegraphics [width=0.325\linewidth]{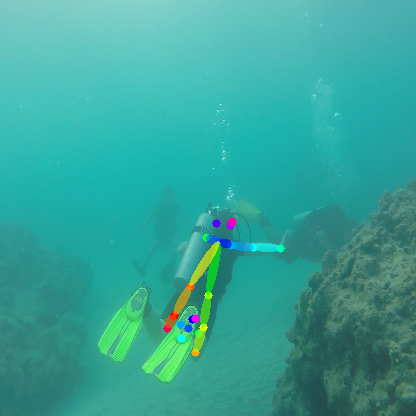} 
	\includegraphics [width=0.325\linewidth]{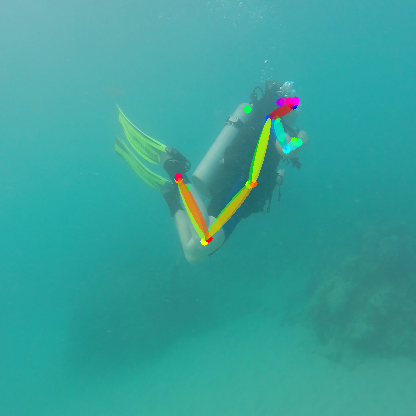} 
	\caption{\footnotesize Multi-human 2D body-pose detection using OpenPose in various human-robot collaborative settings.}%
	\label{fig:openpose}
\end{figure}%

OpenPose generates $18$ key-points pertaining to the nose, neck, shoulders, elbows, wrists, hips, knees, ankles, eyes, and ears of a human body. As shown in Fig.~\ref{fig:openpose}, a subset of these 2D key-points and their pair-wise anatomical relationships are generated for each human. We represent the key-points $\mathbf{KP}(I)$ by a $N_I \times 18$ array where $N_I$ is the number of detected humans in an image $I$. If a particular key-point is occluded or not detected, then the values are left as ($-1$, $-1$). 
We configure $\mathbf{KP}(I)$ in a way that the first row belongs to the left-most person, the second row belongs to the next left-most person, and gradually the last row belongs to the right-most person in the image. This way of sorting the key-points helps to speed up the process of associating the rows of $\mathbf{KP}(I_{leader})$ and $\mathbf{KP}(I_{follower})$. That is, the follower robot needs to make sure that it is pairing the key-points of the \emph{same} individuals. This is important because in practice they might be looking at different individuals, or the same individuals in a different spatial order. Associating multiple persons across different images is a well-studied problem known as \emph{person re-identification (ReId)}.

\begin{figure*}[t]
	\centering
	\includegraphics [width=0.98\linewidth]{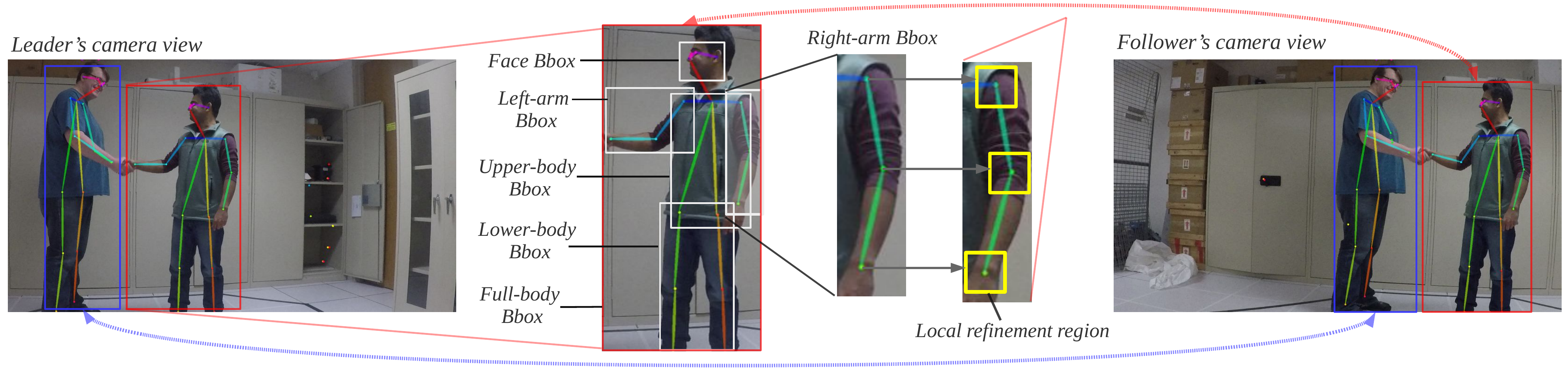}
	\caption{\footnotesize An illustration of how the hierarchical body-parts are extracted for person ReId based on their structural similarities; once the persons are associated, the pair-wise key-points are refined and used as correspondences.}
	\label{fig:association}
\end{figure*}%

\subsection{Person Re-identification using Hierarchical Similarities}
\label{sec:re_id}
Although several existing deep visual models provide very good solutions for person ReId~\cite{ahmed2015improved,li2014deepreid}, we design a simple and efficient model to meet the real-time single-board computational constraints. The idea is to avoid using a computationally demanding feature extractor by making use of the hierarchical anatomical structures that are already embedded in the key-points. First, we bundle the subsets of key-points in several spatial bounding boxes (BBox) as follows:
\begin{itemize}
    \item Face BBox: nose, eyes, and ears;
    \item Upper-body BBox: neck, shoulders, and hips;  
    \item Lower-body BBox: hips, knees, and ankles;
    \item Left-arm BBox: left shoulder, elbow, and wrist; 
    \item Right-arm BBox: right shoulder, elbow, and wrist; 
    \item Full-body BBox: encloses all the key-points. 
\end{itemize}

Fig.~\ref{fig:association} illustrates the spatial hierarchy of these BBoxes and their corresponding key-points. They are extracted by spanning the corresponding key-points' coordinate values in both the $x$ and $y$ dimensions. We use an offset (of additional $10\%$ length) in each dimension to capture more spatial information around the key-points. A BBox is discarded if its area falls below an empirically chosen threshold of $600$ square pixels. We found that BBox areas below this resolution are not always informative and are prone to erroneous results. This happens when the corresponding body-part is either not detected or very far from the camera.

Once the BBox areas are selected, we exploit their pairwise structural properties as features for person ReId; specifically, we compare the structural similarities~\cite{wang2004image} between image patches pertaining to the face, upper-body, lower-body, left-arm, right-arm, and the full body of a person. Based on their aggregated similarities, we evaluate the pair-wise association between each person as seen by the leader (in $I_{leader}$) and by the follower (in $I_{follower}$). 
The structural similarity~\cite{wang2004image} for a particular pair of single-channel rectangular image-patches ($\mathbf{x}$, $\mathbf{y}$) is evaluated based on three properties: luminance $l(\mathbf{x},\mathbf{y}) = {2 \mathbf{\mu}_\mathbf{x} \mathbf{\mu}_\mathbf{y}}/{(\mathbf{\mu}_\mathbf{x}^2+\mathbf{\mu}_\mathbf{y}^2)}$, contrast $c(\mathbf{x},\mathbf{y}) = {2 \mathbf{\sigma}_\mathbf{x} \mathbf{\sigma}_\mathbf{y}}/{(\mathbf{\sigma}_\mathbf{x}^2+\mathbf{\sigma}_\mathbf{y}^2})$, and structure $s(\mathbf{x},\mathbf{y}) = {\mathbf{\sigma}_{\mathbf{xy}}}/{\mathbf{\sigma}_\mathbf{x}\mathbf{\sigma}_\mathbf{y}}$; here, $\mathbf{\mu}_\mathbf{x}$ ($\mathbf{\mu}_\mathbf{y}$) denotes the mean of image patch $\mathbf{x}$ ($\mathbf{y}$), $\mathbf{\sigma}_\mathbf{x}^2$ ($\mathbf{\sigma}_\mathbf{y}^2$) denotes the variance of $\mathbf{x}$ ($\mathbf{y}$), and $\mathbf{\sigma}_{\mathbf{xy}}$ denotes the cross-correlation between $\mathbf{x}$ and $\mathbf{y}$. The structural similarity metric (SSIM) is then defined as: 
\[
    SSIM(\mathbf{x},\mathbf{y}) = l(\mathbf{x},\mathbf{y}) c(\mathbf{x},\mathbf{y})  s(\mathbf{x},\mathbf{y}) = \frac{2 \mathbf{\mu}_\mathbf{x} \mathbf{\mu}_\mathbf{y} }{\mathbf{\mu}_\mathbf{x}^2+\mathbf{\mu}_\mathbf{y}^2} \times \frac{2 \mathbf{\sigma}_{\mathbf{xy}}}{\mathbf{\sigma}_\mathbf{x}^2+\mathbf{\sigma}_\mathbf{y}^2}.     
\]
In order to ensure numeric stability, two standard constants $c_1 = (255k_1)^2$ and $c_2 = (255k_2)^2$ are added as: \begin{equation}
        SSIM(\mathbf{x},\mathbf{y}) =  \frac{2 \mathbf{\mu}_\mathbf{x} \mathbf{\mu}_\mathbf{y} + c_1}{\mathbf{\mu}_\mathbf{x}^2+\mathbf{\mu}_\mathbf{y}^2 + c_1} \times \frac{2 \mathbf{\sigma}_{\mathbf{xy}} + c_2}{\mathbf{\sigma}_\mathbf{x}^2+\mathbf{\sigma}_\mathbf{y}^2 + c_2}. 
\label{eq:one}
\end{equation}
We use $k_1=0.01$, $k_2=0.03$, and an $8\times8$ sliding window in our implementation. Additionally, we resize the patches extracted from $I_{leader}$ so that their corresponding pairs in $I_{follower}$ have the same dimensions. Then, we apply Eq.~\ref{eq:one} on every channel ($R,G,B$) and use their average value as the similarity metric on a scale of [$0$, $1$]. Specifically, we use this metric for person ReId as follows: 
\begin{itemize}
    \item We only consider the mutually visible body-parts for evaluating the pair-wise SSIM values. This choice is important to enforce meaningful comparisons; otherwise, it is equivalent to using only the full-body BBox, which we found to be highly inaccurate.
    \item Each person in $I_{follower}$ is associated with the most similar person corresponding to the maximum SSIM value in $I_{leader}$. However, the association is discarded if that value is less than a threshold $\delta_{min}=0.4$ which is chosen by an AUC (area under the curve)-based analysis (see Section~\ref{impact}). This reduces the risk of inaccurate associations, particularly when there are mutually exclusive people in the scene.           
\end{itemize}

\begin{figure*}[t]
	\centering
	\begin{subfigure}{\textwidth}
		\centering	
		\includegraphics[width=0.92\linewidth]{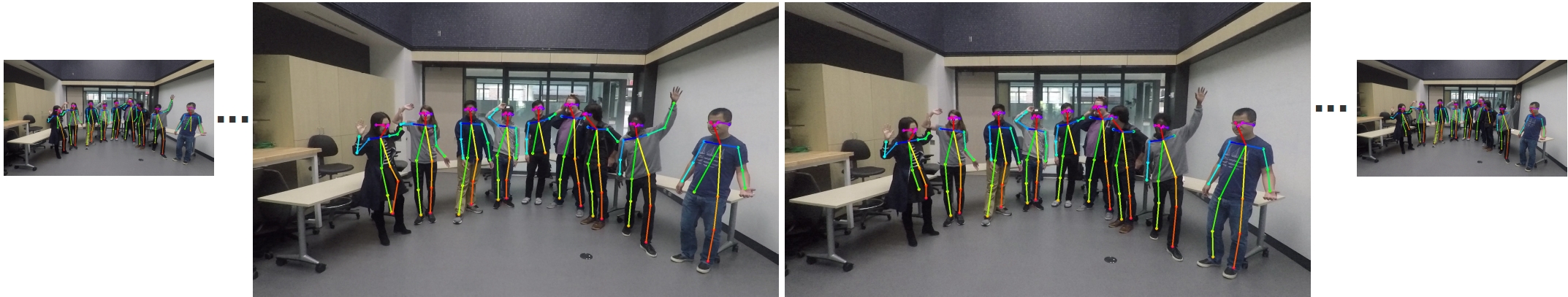}
		\caption{\footnotesize A group of people seen from multiple views and their 2D body-poses (detected by OpenPose).}
		\label{fig:exp1a}
	\end{subfigure}
	\vspace{2mm}
	
	\begin{subfigure}{\textwidth}
		\centering	
		\includegraphics[width=0.86\linewidth]{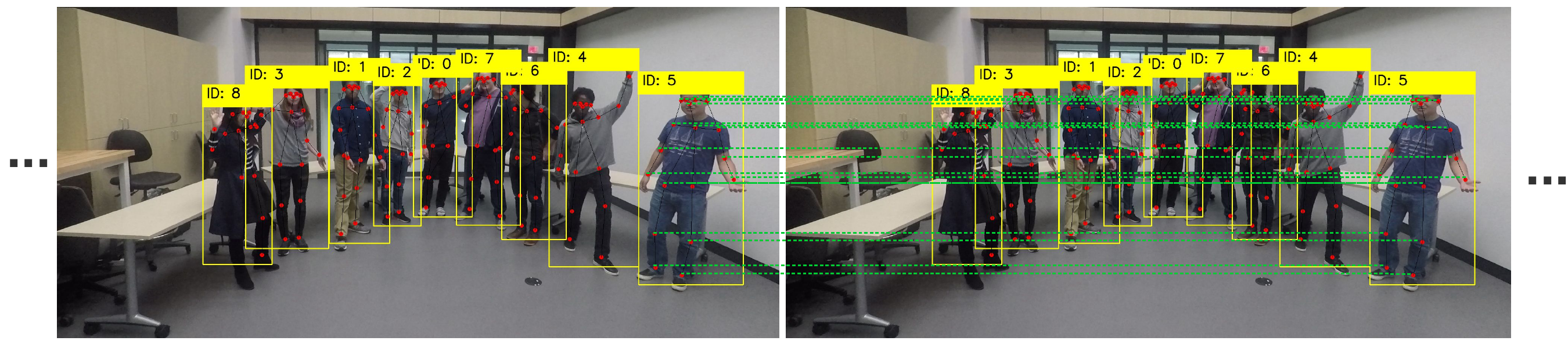}
		\caption{\footnotesize Person association and pose-based key-point correspondences for a particular image pair; a unique identifier is assigned to each association, matched key-points are shown in green lines for the right-most person.}
		\label{fig:exp1b}
	\end{subfigure}
	\vspace{3mm}
	
	\begin{subfigure}{\textwidth}
		\centering	
		\includegraphics[width=0.85\linewidth]{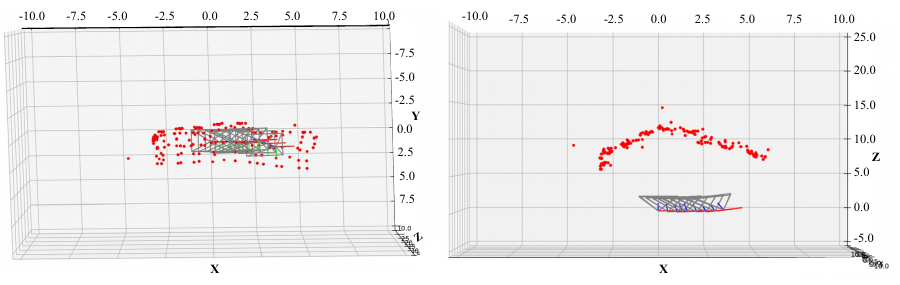}
		\caption{\footnotesize The reconstructed 3D key-points of the humans' structure and the estimated camera poses (up-to scale).}
		\label{fig:exp1c}
	\end{subfigure}
	
	\caption{\footnotesize Results of estimating \emph{structure from motion} using only human pose-based key-points as features.}
	\label{fig:exp1}
\end{figure*}

\begin{figure*}[h]
	\centering
	\begin{subfigure}{0.49\textwidth}
		\centering	
		\includegraphics[width=\linewidth]{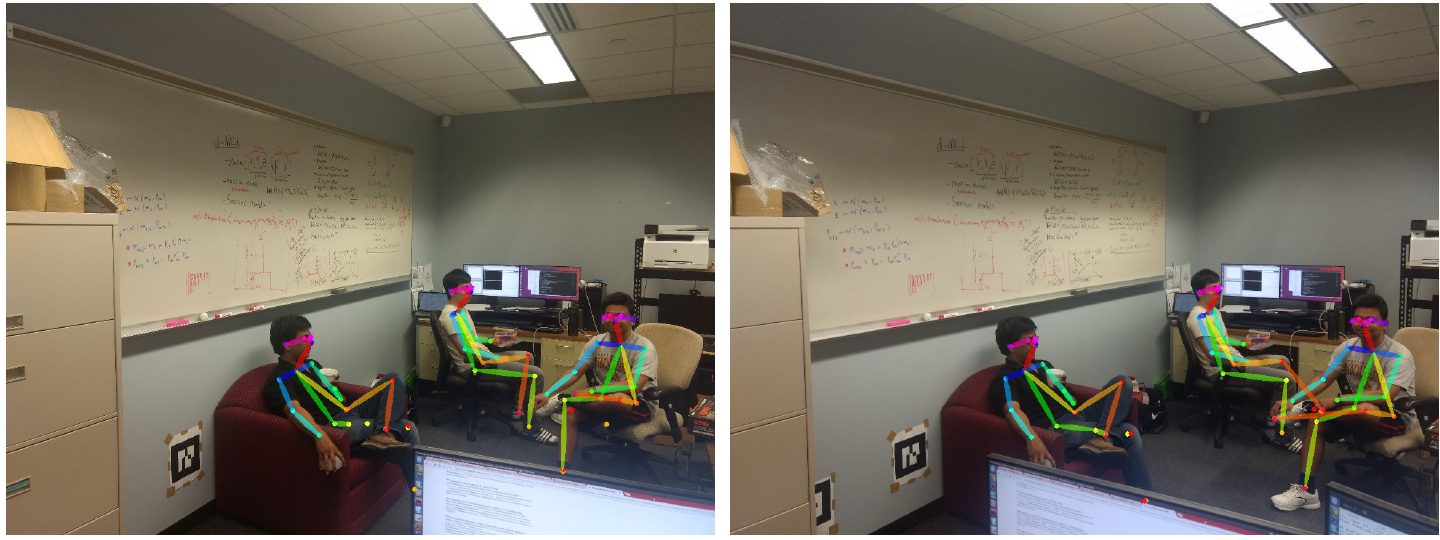}
		\caption{\footnotesize Three humans seen from two different views.}
	\end{subfigure}~
	\begin{subfigure}{0.48\textwidth}
		\centering	
		\includegraphics[width=\linewidth]{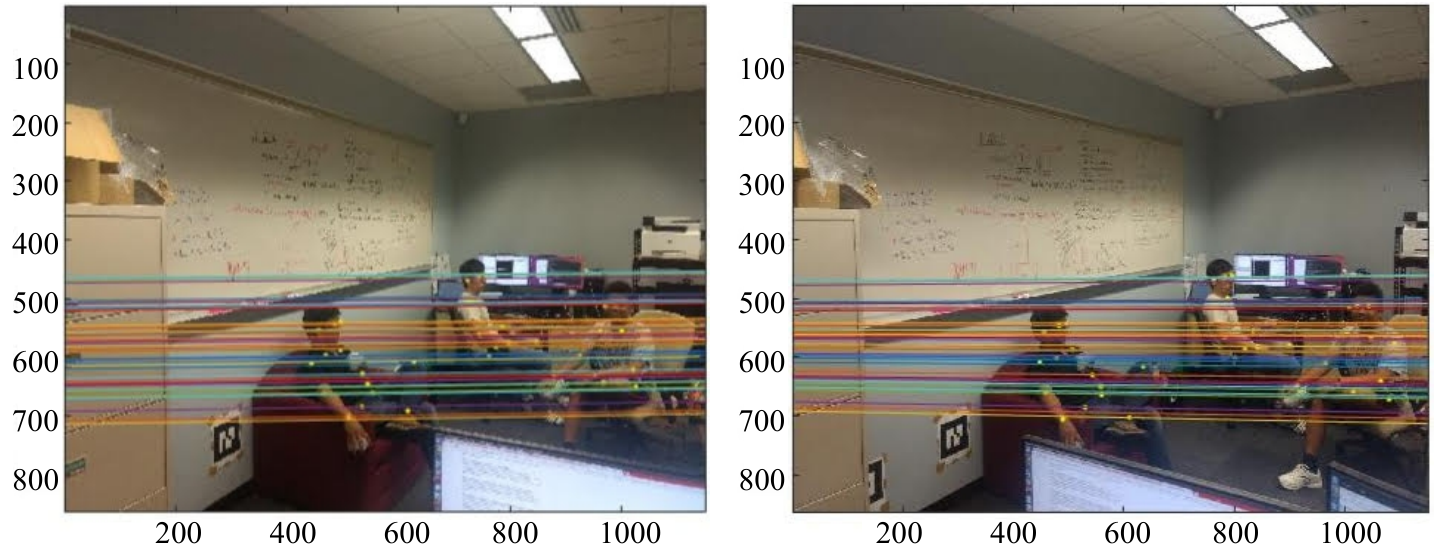}
		\caption{\footnotesize Key-point correspondences and epipolar lines.}
	\end{subfigure}
	\vspace{3mm}
	
	\begin{subfigure}{\textwidth}
		\centering	
		\includegraphics[width=0.9\linewidth]{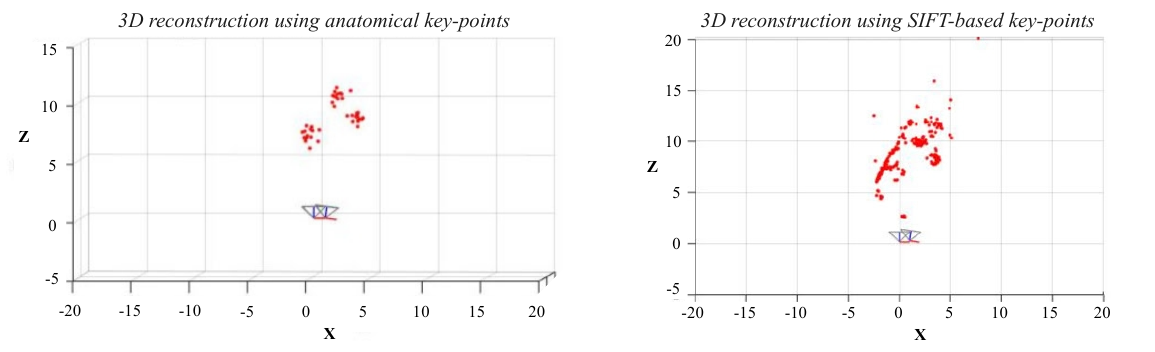}
		\caption{\footnotesize The reconstructed 3D key-points and the estimated camera poses (up-to scale).}
	\end{subfigure}
	\vspace{3mm}
	
	\begin{subfigure}{0.48\textwidth}
		\centering	
		\includegraphics[width=0.8\linewidth]{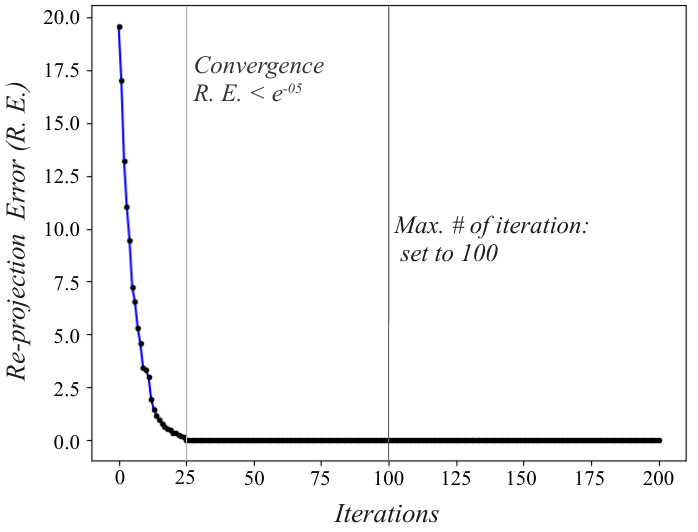}
		\caption{\footnotesize Reduction of the average re-projection error by the iterative key-point refinement process.}
		\label{iterativeness}
	\end{subfigure}~ \hspace{2mm}
	\begin{subfigure}{0.465\textwidth}
		\centering	
		\includegraphics[width=0.8\linewidth]{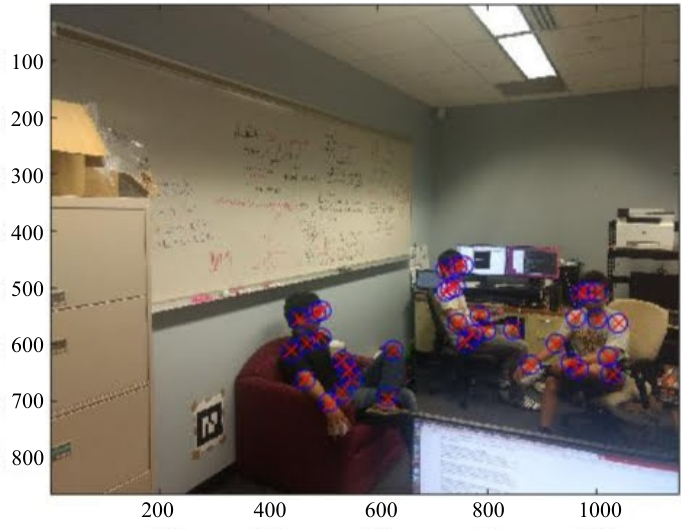}
		\caption{\footnotesize Re-projected points (red crosses) on the left image are shown; the blue circles represent true locations.}
	\end{subfigure}
	
	\caption{Structure from motion for a \emph{two-view} case using only human pose-based key-points as features.}
	\label{fig:appen1}
\end{figure*}

\subsection{Key-point Refinement}
\label{sec:kp_ref}
Once the specific persons are identified, \textit{i.e.}, the rows of $\mathbf{KP}(I_{leader})$ and $\mathbf{KP}(I_{follower})$ are associated, the mutually visible key-points are paired together to form correspondences. Although the key-points are ordered and OpenPose localizes them reasonably well, they cannot be readily used as geometric correspondences due to perspective distortions and noise.  
We attempt to solve this problem by designing an iterative optimization algorithm that refines the noisy correspondences based on their structural properties in a $32\times32$ neighborhood. By denoting $\mathbf{\phi}_I(\mathbf{p})$ as the $32\times32$ image-patch centered at $\mathbf{p}=[p_x, p_y]^T$ in image $I$, we define a loss function for each correspondence $(\mathbf{p}_l \in I_{leader}, \mathbf{p}_f \in I_{follower})$ as:
\begin{equation}
L(\mathbf{p}_l, \mathbf{p}_f) = 1 - SSIM(\mathbf{\phi}_{I_{leader}}(\mathbf{p}_{l}), \mathbf{\phi}_{I_{follower}}(\mathbf{p}_{f})).
\end{equation}
Then, we refine each initial key-point correspondence $(\mathbf{p}_l^{0}, \mathbf{p}_f^{0})$ by minimizing the following function:
\begin{equation}
\mathbf{p}_f^* = \operatorname*{argmin}_{\mathbf{p}} \quad L(\mathbf{p}_{l}^{0}, \mathbf{p}) 
\quad \text{s. t.} \quad ||\mathbf{p}-\mathbf{p}_{f}^{0}||_\infty<32. 
\label{eq:three}
\end{equation}
As Eq.~\ref{eq:three} suggests, we fix $\mathbf{p}_l=\mathbf{p}_l^{0}$ and refine $\mathbf{p}_f=\mathbf{p}_f^{0}$ to maximize $SSIM(\mathbf{\phi}_{I_{leader}}(\mathbf{p}_{l}), \mathbf{\phi}_{I_{follower}}(\mathbf{p}_{f}))$. In our implementation, we adopt a gradient-based refinement algorithm that performs the following iterative update:
\begin{equation}
\mathbf{p}_f^{t+1} = \mathbf{p}_f^t - \eta \cdot \nabla L(\mathbf{p}_l^0, \mathbf{p}_f^t). 
\end{equation}
We follow the procedures suggested in~\cite{avanaki2009exact,otero2014solving} for computing the gradient of SSIM. For fast processing, we vertically stack all the key-points and their gradients to perform the optimization simultaneously with a fixed learning rate of $\eta=0.003$ for a maximum iteration of $100$. We present empirical validations for the choices of the refinement resolution and other hyper-parameters in Section~\ref{impact}.

\subsection{Robot-to-robot Pose Estimation}\label{pnp}
Once the mutually visible key-points are associated and refined, the follower robot uses the corresponding 3D positions (provided by the leader) to estimate its relative pose by solving a PnP problem. Thus, we require that the leader robot is equipped with a stereo camera (or an RGBD camera) so that it can triangulate the refined key-points using epipolar constraints (or use the depth sensor) to represent the key-points in 3D. 

Let $\mathbf{x}_l$ denote the 3D locations of the key-points in the leader's coordinate frame, and $\mathbf{p}_f$ denote their corresponding 2D projections on the follower's camera. Then, assuming the cameras are synchronized, the PnP problem is formulated as follows:
\begin{equation}\label{eq:opt_final}
    \mathbf{T}_f^l = \operatorname*{argmin}_{\mathbf{T}_f^l} ||\mathbf{p}_f - \mathbf{K}_f \mathbf{T}_f^l \mathbf{x}_l||^2.
\end{equation}
Here, $\mathbf{K}_f$ is the intrinsic matrix of the follower's camera and $\mathbf{T}_f^l$ is its 6-DOF transformation relative to the leader. In our implementation, we follow the standard iterative solution for PnP using RANSAC~\cite{zheng2013revisiting}.

\section{Experimental Analysis}
We conduct several experiments with 2-DOF and 3-DOF robots to evaluate the performance and feasibility of the proposed relative pose estimation method. We present these experimental details, analyze the results, and discuss various operational considerations in the following sections.

\subsection{Proof of Concept: Structure from Motion}
At first, we perform experiments to validate that the human pose-based key-points can be used as geometric correspondences for relative pose estimation. As illustrated in Fig.~\ref{fig:exp1a}, we emulate an experimental set-up for \emph{structure from motion} with humans; we use an intrinsically calibrated monocular camera to capture a group of nine (static) people from multiple views. Here, the goal is to estimate the camera poses and reconstruct the 3D structures of the humans using only their body-poses as features. 

In the evaluation, we first use OpenPose to detect the human pose-based 2D key-points in the images (Fig.~\ref{fig:exp1a}). Then, we utilize the proposed person ReId and key-point refinement modules to obtain the feature correspondences across multiple views (Fig.~\ref{fig:exp1b}). Subsequently, we follow the standard procedures for structure from motion~\cite{hartley2003multiple}: fundamental matrix computation using 8-point algorithm with RANSAC, essential matrix computation, camera pose estimation by enforcing the Cheirality constraint, and linear triangulation. Finally, the triangulated 3D points and camera poses are refined using bundle adjustment. As demonstrated in Fig.~\ref{fig:exp1c}, the spatial structure of the reconstructed points on the human bodies and the camera poses are consistent with our setup. Results of another experiment for a \emph{two-view} case are shown in Fig.~\ref{fig:appen1}, which further validate that the estimated camera poses are comparable to the ground truth, \ie, analogous SIFT feature-based estimation. In the next section, we demonstrate the effectiveness of our proposed body-pose association and key-point refinement modules in ensuring this robust pose estimation performance.

\begin{table*}[t]
	\centering
	\footnotesize
	\caption{A quantitative performance comparison for various person ReId models on standard datasets; a set $150$ test images are used for comparison from each dataset.}
	\begin{tabular}{l||c|c||c|c||c}
		\cline{1-1} \cline{2-6}
		Person ReId & \multicolumn{2}{c||}{{\tt Market-1501 Dataset}} & \multicolumn{2}{c||}{{\tt CUHK-03 Dataset}} & {FPS} \\ \cline{2-5}
		models &  Rank-1 Acc. ($\%$) & MAP ($\%$) & Rank-1 Acc. ($\%$) & MAP ($\%$) & (Jetson TX2) \\
		\hline 
		Aligned ReId  & $92.90$ & $90.12$  &  $67.15$ & $68.03$ & $0.67$ \\
		Deep person ReId  & $85.86$ & $68.24$      & $62.26$ & $65.15$ & $0.33$ \\
		Tripled-loss ReId  & $85.25$ & $74.88$   & $72.75$ & $60.27$  & $0.62$  \\
		\textbf{Proposed person ReId}  & $75.67$ & $72.26$   & $57.82$ & $54.91$  & \textbf{7.45}  \\
		\hline 
	\end{tabular}
	\label{tab:std}
	\vspace{3mm}
	
	\caption{Effectiveness of the proposed person ReId method on real-world data; each set contains $100$ images of multiple humans in ground and underwater scenarios. }
	\begin{tabular}{l||c|c||c|c}
		\cline{1-1} \cline{2-5}
		Person ReId & \multicolumn{2}{c||}{{\tt Set A (1-2 humans per image)}} & \multicolumn{2}{c}{{\tt Set B (3-5 humans per image)}} \\ \cline{2-5}
		models &  Rank-1 Acc. ($\%$) & FPS (Jetson TX2) & Rank-1 Acc. ($\%$) & FPS (Jetson TX2)  \\
		\hline 
		Aligned ReId  & $62.75$ & $0.62$  &  $56.65$ & $0.48$  \\
		Deep person ReId   & $55.32$ & $0.29$      & $42.36$ & $0.12$  \\
		Tripled-loss ReId  & $55.15$ & $0.58$   & $44.85$ & $0.44$  \\
		\textbf{Proposed person ReId}  & \textbf{76.55} & \textbf{6.81}   & \textbf{71.56} & \textbf{5.45}    \\
		\hline 
	\end{tabular}
	\label{tab:our}
\end{table*}

\begin{figure*}[h]
	\centering
	\vspace{1mm}
	\begin{subfigure}{0.46\textwidth}
		\centering	
		\includegraphics[width=0.9\linewidth]{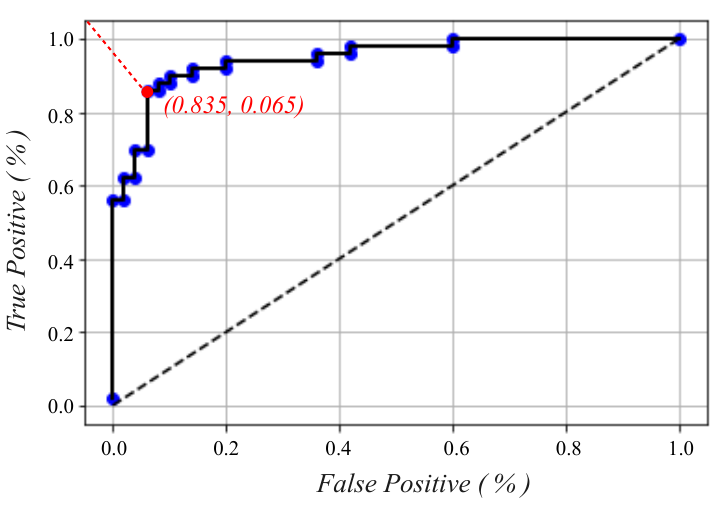}
		\caption{\footnotesize ROC curve for the person ReID performance (AUC $= 0.938$): a total of $20$ thresholds ($\delta_{min} \in [0,1]$) are considered in the evaluation; the point marked in red corresponds to $\delta_{min}=0.4$.}
		\label{fig:param_a}
	\end{subfigure}~ \hspace{4mm}
	\begin{subfigure}{0.46\textwidth}
		\centering	
		\includegraphics [width=0.9\linewidth]{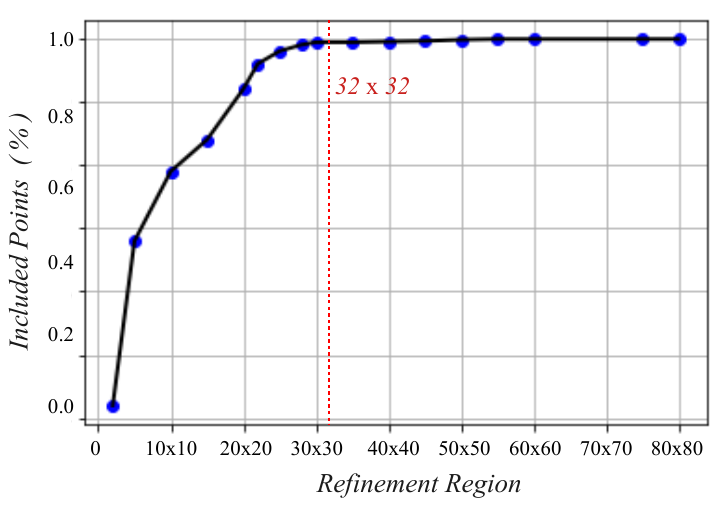}
		\caption{\footnotesize Fraction of key-point correspondences that fall within various resolutions of respective refinement regions; the vertical red line corresponds to a refinement region of $32\times32$.}
		\label{fig:param_b}
	\end{subfigure}
	\caption{\footnotesize Empirical selection of hyper-parameters: (a) SSIM threshold for pose association in the proposed person ReId module and (b) resolution of the key-point refinement region. The evaluation is performed on the combined set of $250$ images containing a total of $687$ person associations with $8256$ key-point correspondences.}
	\label{fig:hyper_param}
\end{figure*}

\subsection{Effectiveness of the Body-pose Association and Key-point Refinement Modules}\label{impact}
It is easy to notice that person ReId is essential for associating mutually visible persons across different views. As mentioned in Section~\ref{sec:re_id}, we focus on achieving fast association by making use of the local structural properties around the anatomical key-points in the image space. In contrast, the SOTA person ReId approaches adopt deep visual feature extractors that are computationally demanding. In Table~\ref{tab:std}, we quantitatively evaluate the SOTA models named Aligned ReId~\cite{zhao2017deeply}, Deep person ReId~\cite{li2014deepreid}, and Tripled-loss ReId~\cite{zheng2011person} based on accuracy and mean averaged precision (mAP) on two standard datasets. Specifically, a test-set containing $150$ instances from the {\tt Market-1501} and {\tt CUHK-03} datasets are used for the evaluation; also, their run-times on a NVIDIA\texttrademark{} Jetson TX2 are shown for comparison. The results indicate that although these models (once trained on similar data) perform well on standard datasets, they are computationally too expensive for single-board embedded platforms.

\begin{figure*}[t]
\centering
	\begin{subfigure}{0.475\textwidth}
		\centering	
		\includegraphics [width=0.9\linewidth]{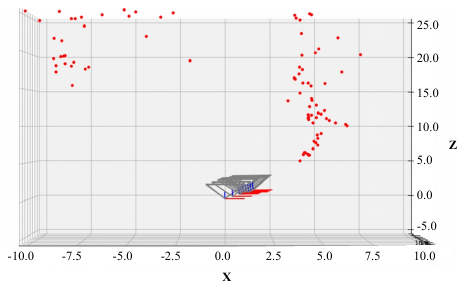}
		\caption{\footnotesize Inaccurate 3D reconstruction using raw key-point correspondences (without refinement).}
		\label{fig:issues_b}
	\end{subfigure}~ \hspace{1mm}
	\begin{subfigure}{0.465\textwidth}
		\centering	
		\includegraphics[width=0.9\linewidth]{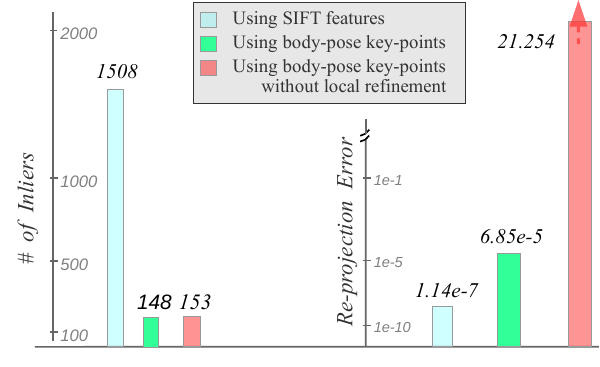}
		\caption{\footnotesize Quantitative performance compared to using SIFT-based features.}
		\label{fig:issues_a}
	\end{subfigure}
	\caption{\footnotesize Necessity of the proposed key-point refinement process; results correspond to the experiment illustrated in Figure~\ref{fig:exp1}.}
	\label{fig:issues}
\end{figure*}

Moreover, as demonstrated in Table~\ref{tab:our}, these off-the-shelf models do not perform that well on high-resolution real-world images. Although their performance can be improved by training on more comprehensive real-world data, the computational complexity remains a barrier. To this end, the proposed person Reid module provides significantly faster run-time and better portability as it does not require rigorous large-scale training. Its only hyper-parameter is the SSIM threshold $\delta_{min}$ (see Section~\ref{sec:re_id}), which we select by standard AUC-based analysis of ROC (receiver operating characteristic) curve. As shown in Figure~\ref{fig:param_a}, we choose $\delta_{min}=0.4$, which corresponds to $83.5\%$ true-positive and $6.5\%$ false-positive rates for person ReID on the combined test set of $250$ images containing $687$ person associations. Additionally, we select the key-point refinement resolution through an ablation experiment with $8256$ key-point correspondences. We observe that the optimal key-point location is found within $25\times25$ pixels of the initial estimate by OpenPose for over $96\%$ of the cases. As shown in Figure~\ref{fig:param_b}, we make a more conservative choice of $32\times32$ refinement region in our implementation.

Finally, we evaluate the utility and effectiveness of the proposed key-point refinement algorithm based on re-projection errors and compare the results with traditional SIFT feature-based reconstruction. 
As Figure~\ref{fig:issues_b} demonstrates, the 3D reconstruction and camera pose estimation with raw key-points are inaccurate as the unrefined correspondences are invalid in a perspective geometric sense. 
As Figure~\ref{fig:issues_a} shows, the average re-projection error for the refined key-points reduces to $6.85e^{-5}$ pixels, which is acceptable considering the fact that there are ten times less anatomical key-points than SIFT feature-based key-points. This evaluation corresponds to the experiment presented in Figure~\ref{fig:exp1}, which shows that the refined key-points constitute accurate scene reconstruction and camera pose estimation. Another qualitative validation of the iterative key-point refinement algorithm and its convergence behavior can be found in Figure~\ref{fig:appen1}.

\subsection{Robot-to-robot 3-DOF Pose Estimation}
\begin{figure}[h]
\centering
\includegraphics [width=0.98\linewidth]{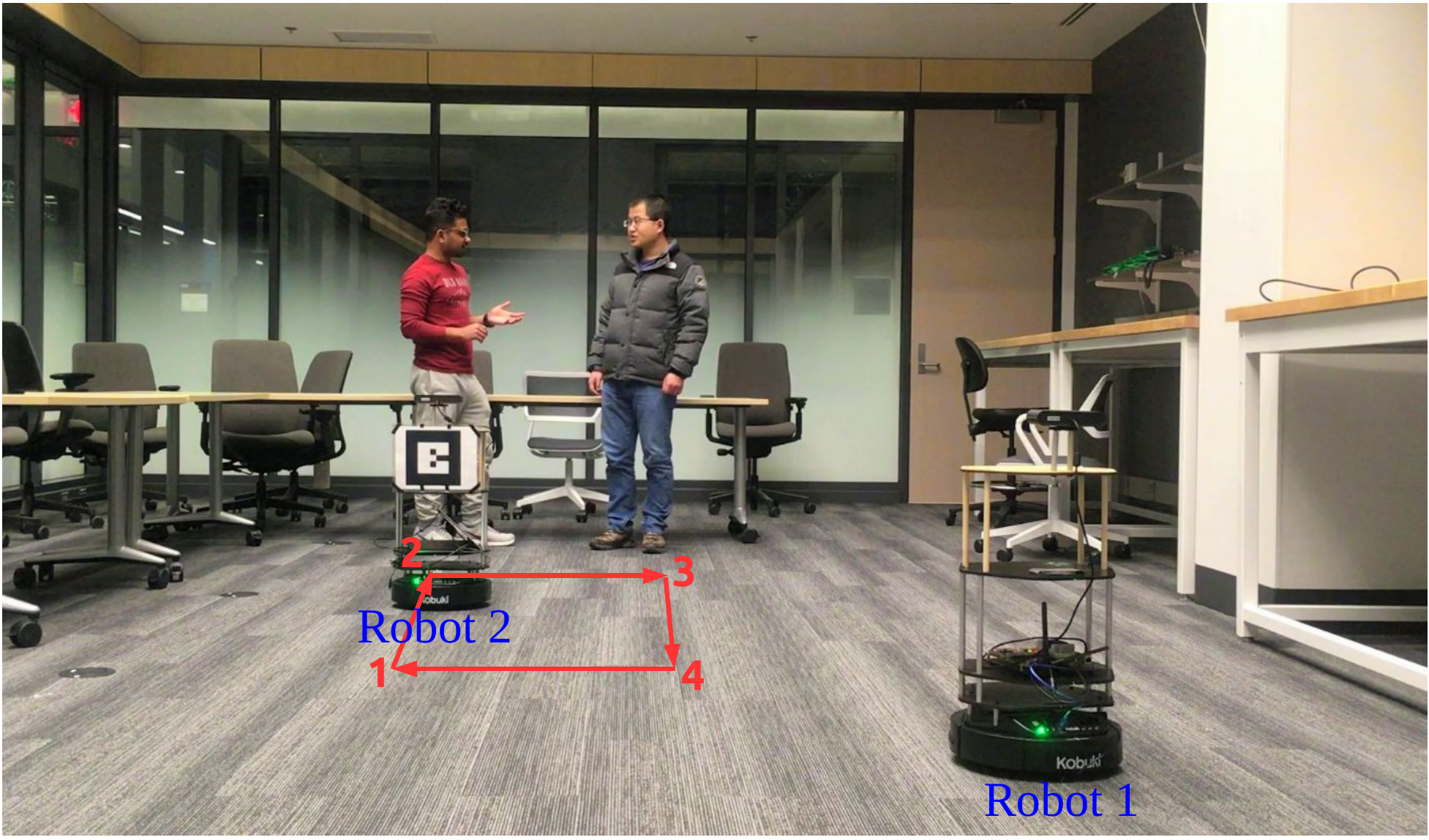}
\caption{\footnotesize 3-DOF ground experiment: one leader and one follower robot; the follower robot's trajectory is shown by red arrows.}
\label{setup_a}
\end{figure}

We also perform experiments for 3-DOF robot-to-robot relative pose estimation with 2D robots.
In the particular scenario shown in Figure~\ref{setup_a}, we use two planar robots (one leader and one follower) and two mutually visible humans in the scene. The robot with an AR-tag on its back is used as the follower robot while the other robot is used as the leader. The AR-tag is used to obtain the follower's ground truth relative pose for comparison. On the other hand, the leader robot is equipped with an RGBD camera; it communicates with the follower and shares the 3D locations of the mutually visible key-points. Specifically, it detects the human pose-based 2D key-points and associates the corresponding depth information to represent them in 3D. Subsequently, the follower robot uses this information to localize itself relative to the leader by following the proposed estimation method.

\begin{figure*}[t]
	\centering
	\begin{subfigure}[t]{0.4\textwidth}
		\centering	
        \includegraphics [width=0.85\linewidth]{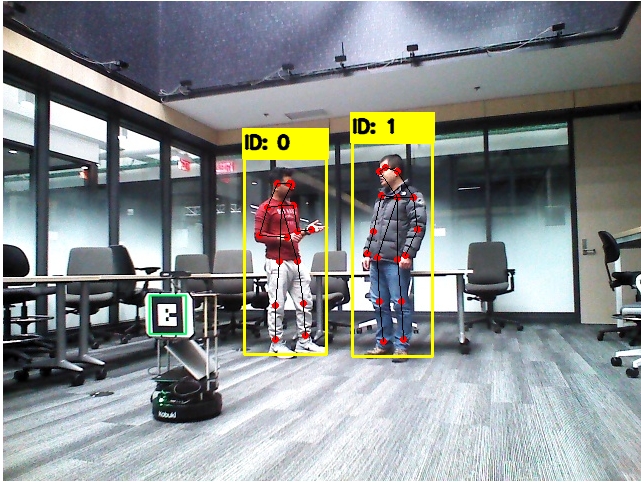}
		\caption{\footnotesize The leader robot detects the pose-based key-points and shares the 3D locations.}
		\label{fig:ground_a}
	\end{subfigure}~ \hspace{2mm}
	\begin{subfigure}[t]{0.5\textwidth}
	\centering	
        \includegraphics [width=0.85\linewidth]{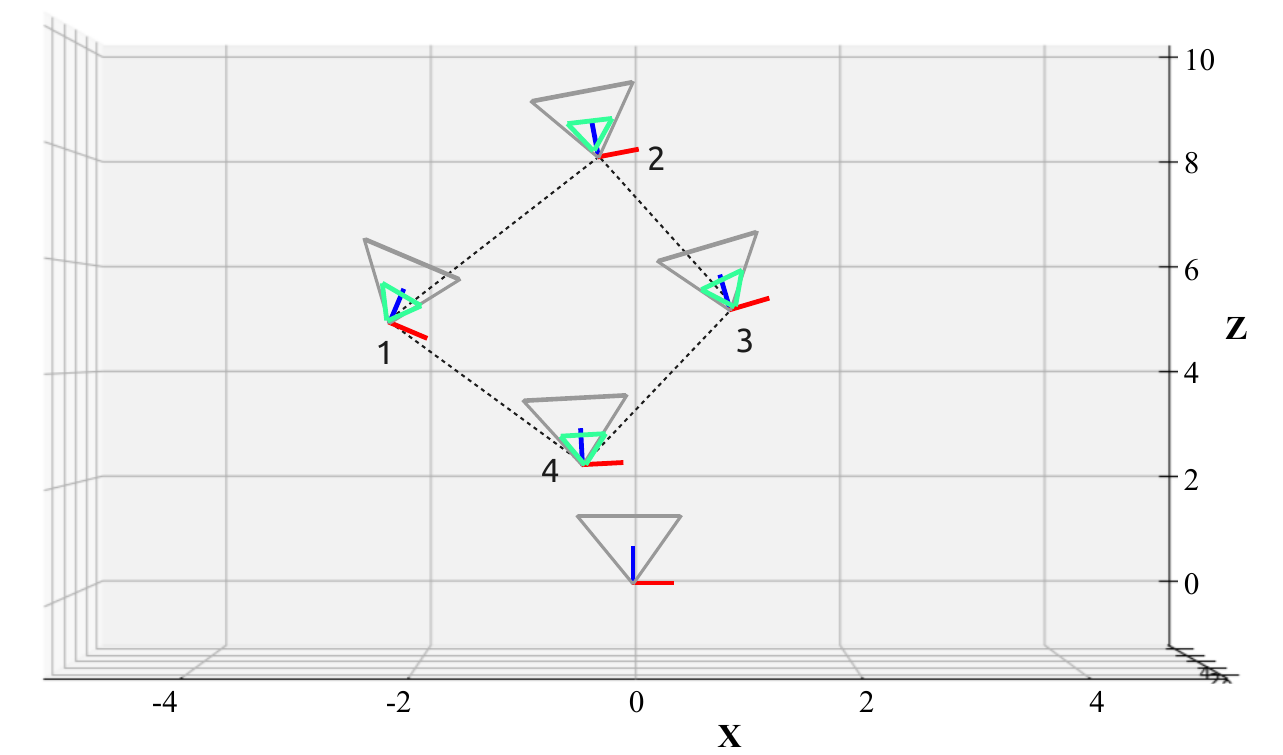}
	\caption{\footnotesize Estimated poses of the follower relative to the leader (the green cones represent the respective ground truth).}
	\label{fig:ground_b}
	\end{subfigure}
    \caption{\footnotesize An experiment to evaluate the accuracy of 2D relative pose estimation with two planar robots and two mutually visible humans.}
\label{fig:ground}
\end{figure*}

\begin{figure*}[t]
    \vspace{1mm}
	\centering
	\begin{subfigure}[t]{\textwidth}
		\centering	
        \includegraphics[width=0.98\linewidth]{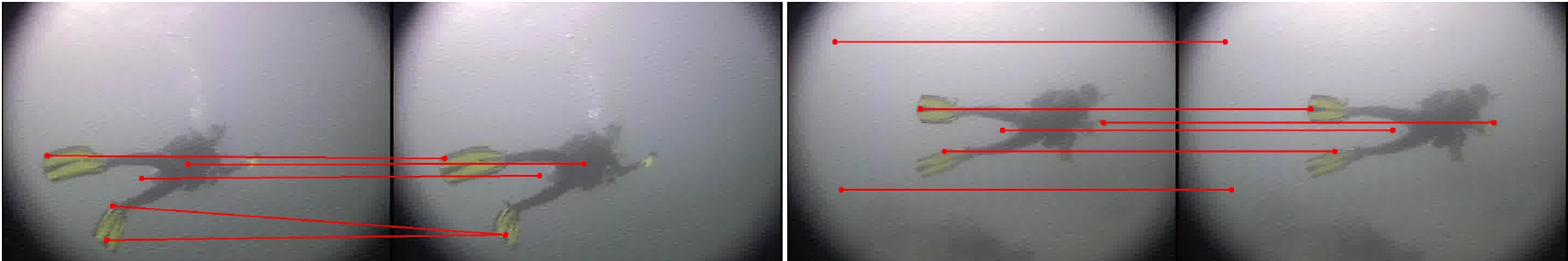} 
		\caption{\footnotesize SIFT feature correspondences are shown for two image pairs (standard FLANN package~\cite{FLANN} is used for matching).}
		\label{fig:exp20a}
	\end{subfigure}
	\vspace{2mm}
	
		\begin{subfigure}[t]{\textwidth}
		\centering	
        \includegraphics[width=0.98\linewidth]{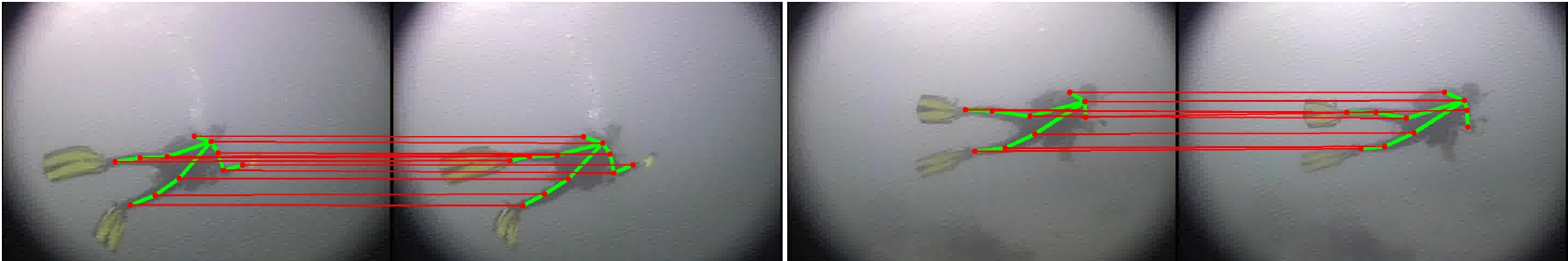} 
		\caption{\footnotesize Human pose-based key-point correspondences are shown for the same image pairs as (a).}
		\label{fig:exp20b}
	\end{subfigure}
	
\caption{\footnotesize Illustrations where a lack of natural landmarks limits the utility of standard feature detectors. As seen, presence of a single human in the scene facilitates considerably more anatomical key-point correspondences than the point-based features. 
}
\label{fig:exp20}
\end{figure*}

\begin{figure}[h]
	\centering
	\includegraphics [width=0.98\linewidth]{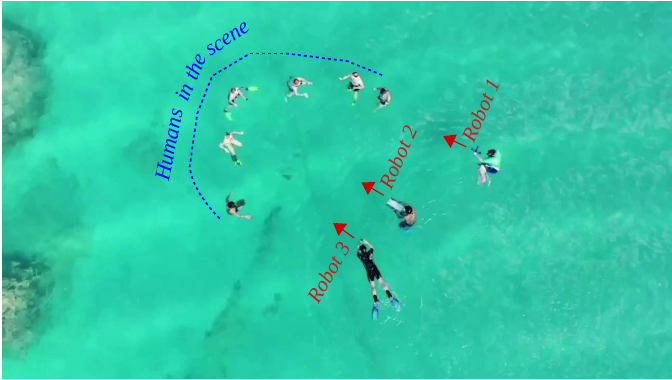}
	\caption{\footnotesize 6-DOF underwater experiment: one leader and two follower robots (aerial view).}
	\label{setup_b}
\end{figure}%

As demonstrated in Figure~\ref{setup_a}, we move the follower robot in a rectangular pattern and evaluate the 3-DOF pose estimates relative to the static leader robot. We present the qualitative results in Figure~\ref{fig:ground}; it shows that the follower robot's pose estimates are very close to their respective ground truth. Overall, we observe an average error of $0.0475\%$ in translation (cm) and a $0.8625^{\circ}$ average error in rotation, which is reasonably accurate. We obtain similar qualitative and quantitative performance with a dynamic leader as well. Next, we present field experimental validations of the relative pose estimation performance in feature-deprived underwater scenarios.

\begin{figure*}[t]
	\centering
	\begin{subfigure}[t]{\textwidth}
		\centering	
        \includegraphics[width=0.98\linewidth]{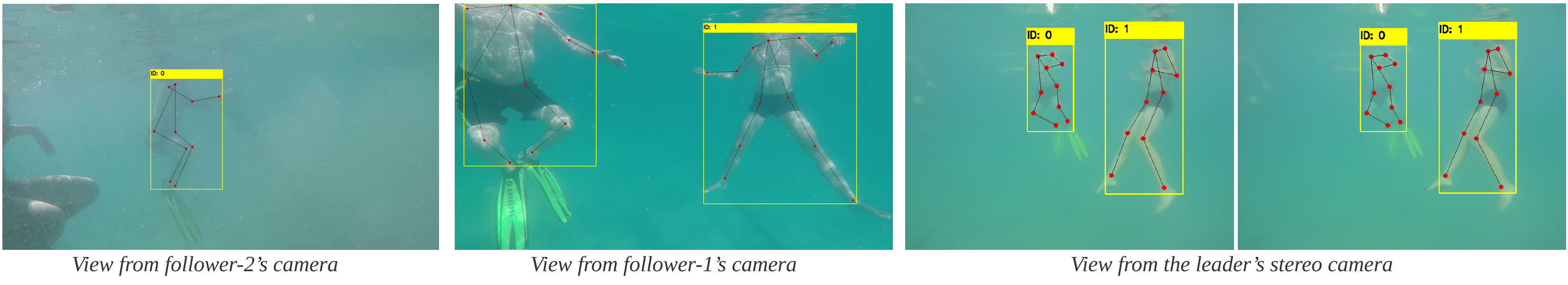} 
		\caption{\footnotesize A group of people seen from multiple perspectives; the detected key-points and their associations are annotated in respective images.}
		\label{fig:exp2a}
	\end{subfigure}
	\vspace{2mm}

	\begin{subfigure}[t]{\textwidth}
		\centering	
        \includegraphics[width=0.92\linewidth]{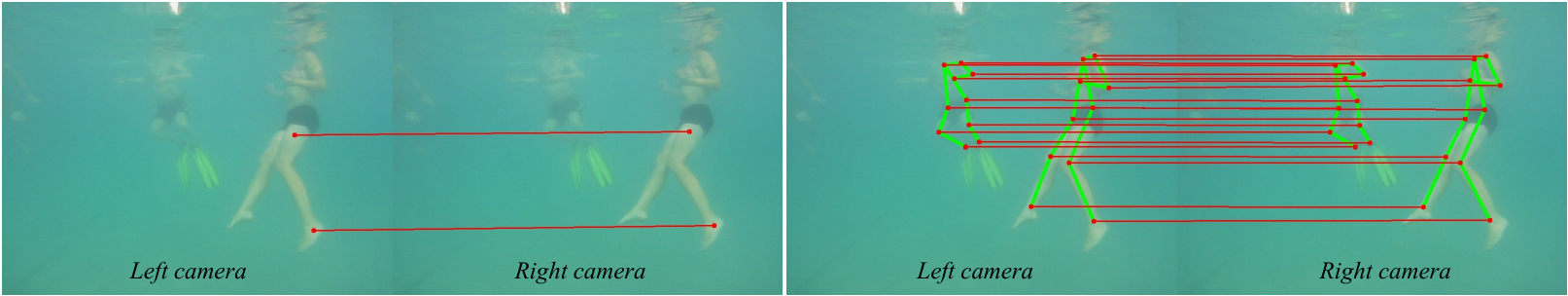} 
		\caption{\footnotesize Feature correspondences for the leader's stereo image pair is shown: (left) SIFT features; (right) human pose-based key-points. As seen, $20$ anatomical key-point matches are found compared to only two SIFT feature matches.}
		\label{fig:exp20c}
	\end{subfigure}
	\vspace{2mm}
	
	\begin{subfigure}[t]{\textwidth}
		\centering	
        \includegraphics[width=0.92\linewidth]{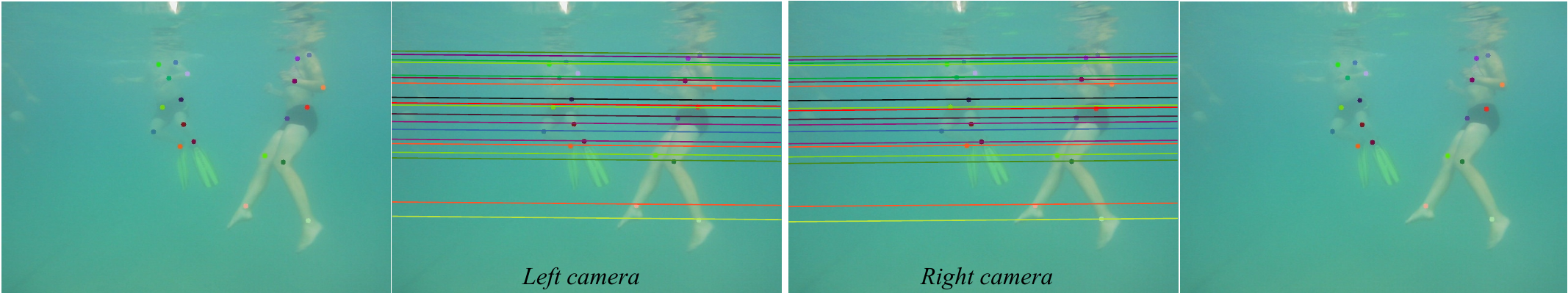} 
		\caption{\footnotesize The anatomical key-points are pair-wise associated, refined, and then used to project epipolar lines for the stereo image pair. }
		\label{fig:exp20e}
	\end{subfigure}
	\vspace{2mm}

	\begin{subfigure}[t]{0.62\textwidth}
		\centering	
        \includegraphics[width=0.49\linewidth]{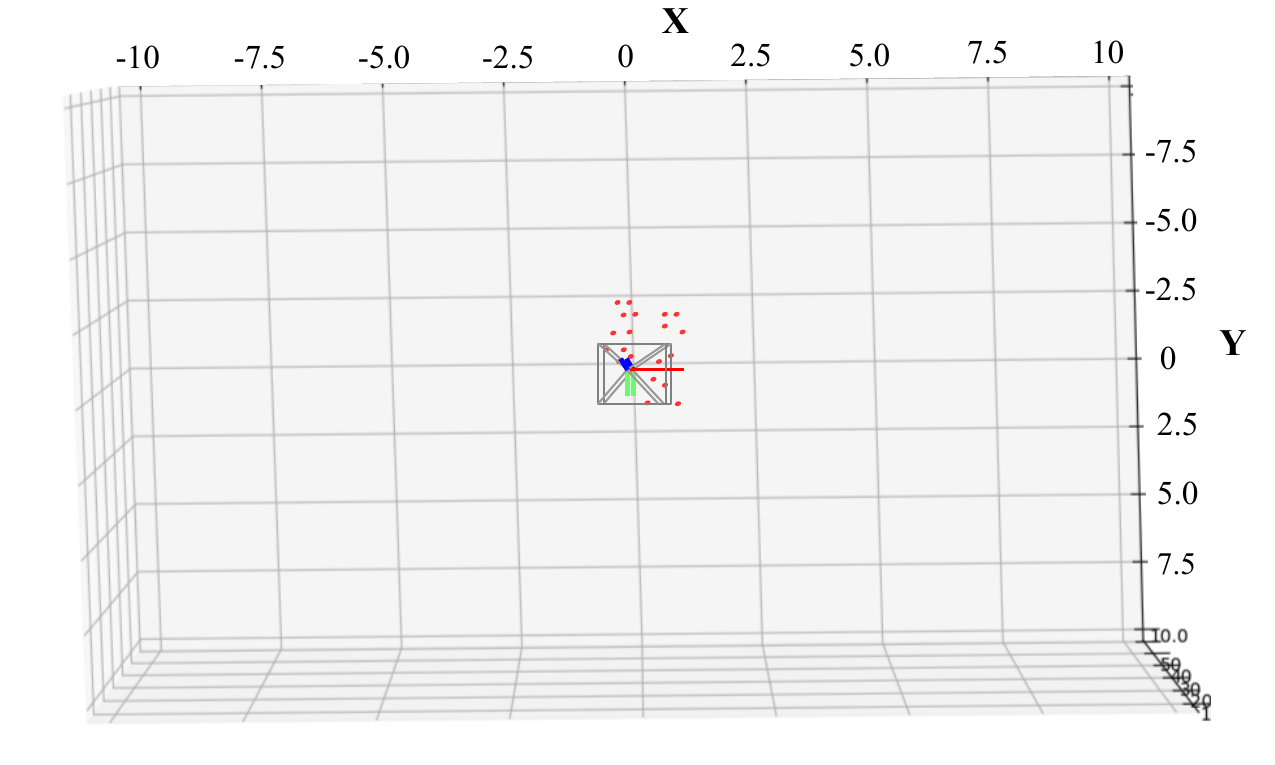}~
		\includegraphics[width=0.49\linewidth]{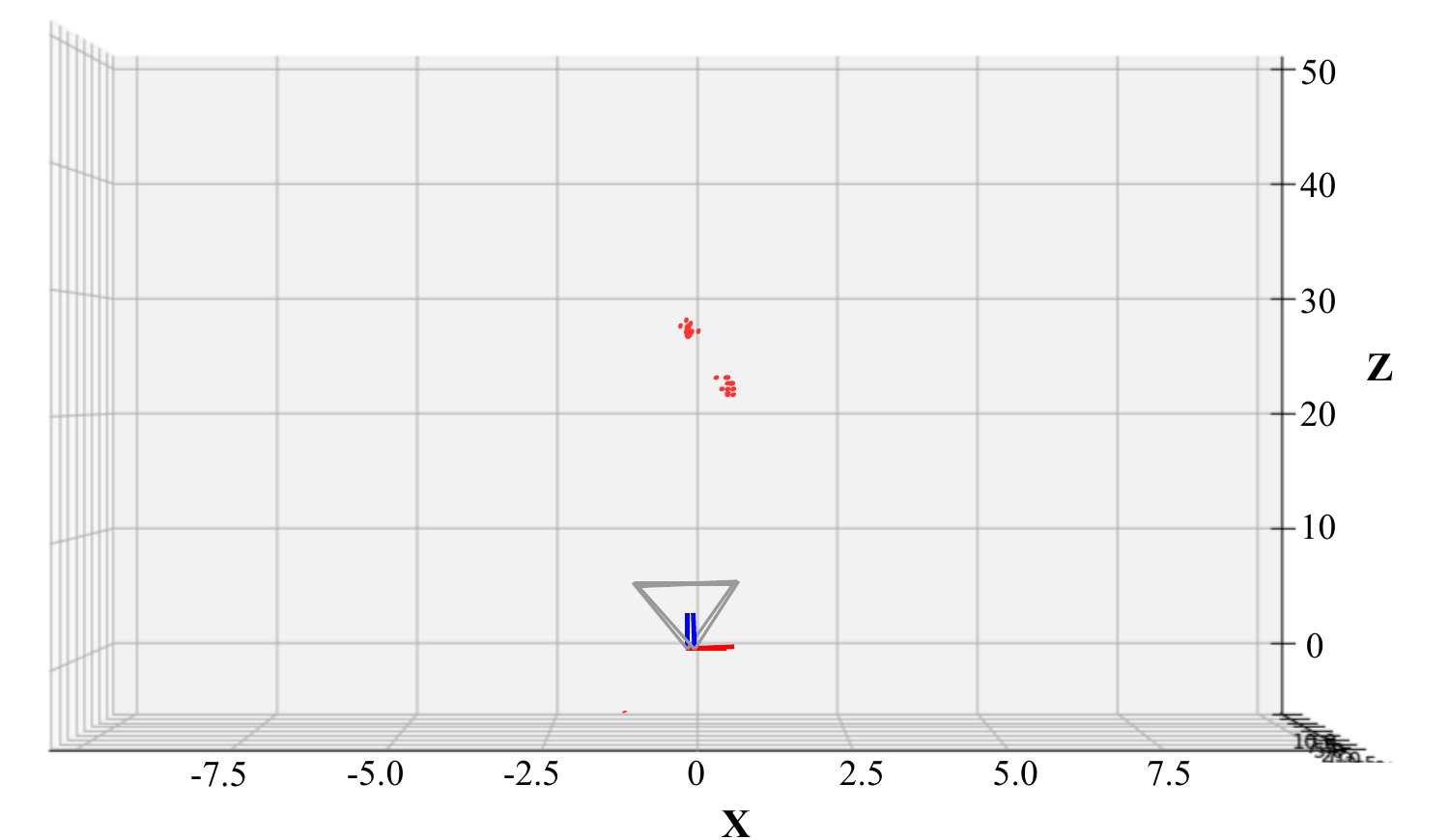}
		\caption{\footnotesize Stereo triangulation of the human pose-based key-points (seen by the leader robot).}
		\label{fig:exp2b}
	\end{subfigure}~
	\begin{subfigure}[t]{0.35\textwidth}
		\centering	
        \includegraphics[width=0.88\linewidth]{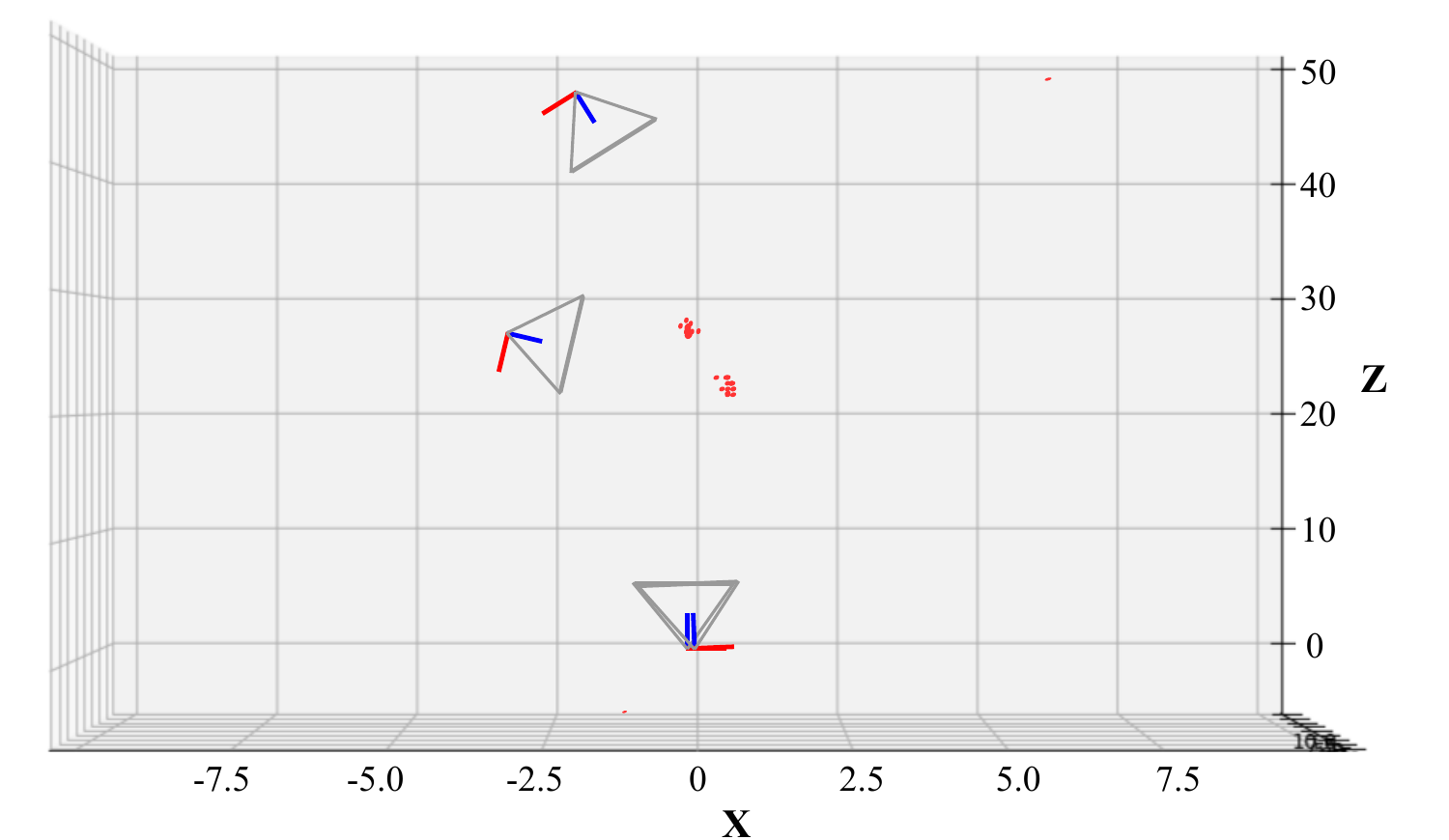}
		\caption{\footnotesize Estimated relative poses of the followers robots.}
		\label{fig:exp2c}
	\end{subfigure}
	
\caption{\footnotesize An underwater experiment for 3D relative pose estimation using one leader and two follower robots.}
\label{fig:exp2}
\end{figure*}

\begin{figure*}[t]
\centering
\includegraphics [width=0.95\linewidth]{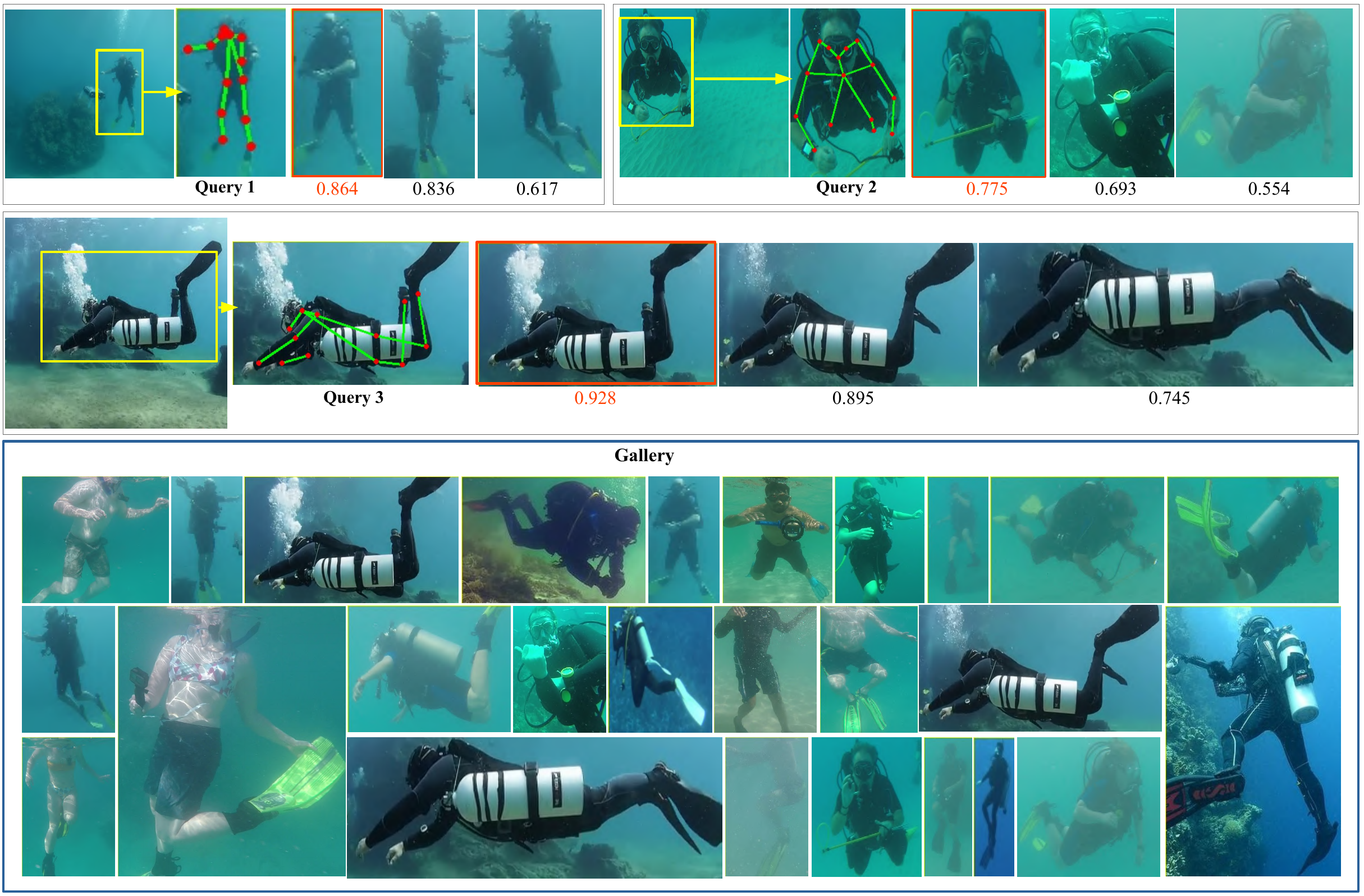}%
\vspace{-1mm}
\caption{\footnotesize Three test cases for the proposed person ReId module are shown: each query is matched with a gallery of candidate images (inside the blue box); the top three matches and respective scores are shown alongside the query image. The scores represent averaged SSIM scores for the mutually visible body-part BBoxes (see Section~\ref{sec:re_id}).     
}
\label{reid}
\end{figure*}%

\subsection{Robot-to-robot 6-DOF Pose Estimation in Adverse Underwater Visual Conditions} 
As seen in Figure~\ref{fig:exp20}, standard point-based feature detectors fail to generate a large pool of reliable correspondences when there are very few salient features and landmarks in the scene. Consequently, the sampling-based parameter estimation techniques (\eg,~RANSAC) often generate inaccurate results in feature-deprived underwater scenarios. However, we demonstrate that human pose-based key-points can still be refined to establish reliable geometric correspondences for robot-to-robot relative pose estimation. Moreover, we get a reasonably large pool of correspondences with only one or two humans in the scene, which is fairly common in cooperative underwater missions.

We perform several field experiments in human-robot collaborative setups; Figure~\ref{setup_b} shows the setup of a particular underwater experiment where we capture human body-poses from different perspectives to estimate the 6-DOF transformations of two follower robots relative to a leader robot. The leader robot is equipped with a stereo camera; hence, the 3D information of the human pose-based key-points is obtained by using stereo triangulation technique. Subsequently, we find the corresponding 2D projections on the follower robots' cameras using the proposed person ReId and key-point refinement processes. Finally, we estimate the follower-to-leader relative poses from their respective PnP solutions.

We present a particular snapshot in Figure~\ref{fig:exp2a}; it illustrates the leader and follower robots' perspectives and the associated human pose-based key-points. 
Subsequently, Figures~\ref{fig:exp20c}-\ref{fig:exp20e} demonstrate the geometric validity of those key-point correspondences and the reconstructed 3D points are shown in Figure~\ref{fig:exp2b}.   As seen, the estimated 3D structure is consistent with the mutually visible humans' body-poses. Finally, the estimated 6-DOF poses of the follower robots relative to the leader robot are shown in Figure~\ref{fig:exp2c}.

Such leader-to-follower pose estimates are useful in cooperative diver following~\cite{islam2018person}, convoying~\cite{shkurti2017underwater}, and other interactive tasks while operating in close proximity. The robust performance and efficient implementation of the proposed modules make it suitable for use by visually-guided underwater robots in human-robot collaborative applications. However, there are a few practicalities involved which can affect the performance; next, we discuss these aspects and their possible solutions based on our experimental findings.

\vspace{3mm}
\subsection{Discussion: Operational Challenges and Practicalities}\label{issues}
\textbf{Synchronized cooperation}: A major operational requirement of multi-robot cooperative systems is the ability to register synchronized measurements in a common frame of reference, which can be quite challenging in practice. For problems such as ours, an effective solution is to maintain a buffer of time-stamped measurements and register those as a batch using a temporal \textit{sliding window}. 
We effectively used such timestamp-based buffer schedulers~\cite{RosTime} in our implementation with reasonable robustness. However, the challenge remains in finding instantaneous relative poses, especially when both robots are in motion. Nevertheless, these aspects are independent of the choice of features/key-points for relative pose estimation and more generic requirements to multi-robot cooperative systems.

\vspace{1mm}
\textbf{Trade-off between robustness and efficiency}: It is quite challenging to ensure a fast yet robust performance for visual feature-based body-pose estimation and person ReId on limited computational resources of embedded platforms. This trade-off between robustness and efficiency led us to design fast body-pose association and refinement modules. These efficient modules enable us to achieve an average end-to-end run-time of $375$-$420$ milliseconds for relative pose estimation on Jetson TX2. Note that the proposed person ReId and key-point refinement account for only $195$-$240$ milliseconds (up to nine humans in the scene). Hence, faster human body-pose detectors (than OpenPose) can significantly boost the end-to-end run-time of the system.

In Section~\ref{impact}, we demonstrated that the proposed person ReId model performs reasonably well in practice despite its simplistic design. One operational benefit in our application is that the humans are seen at once from every perspective; hence, both their appearances and body-poses remain consistent. We provide a demonstration of this benefit in Figure~\ref{reid}; it shows three \emph{queries} for ReId where humans with similar suit/wearable are seen at various distances and orientations from the camera. Although the \emph{gallery} images contain several humans with similar appearances, we observe that the top three results correspond to best matches both in terms of human appearance and body-pose. We postulate that computing aggregated similarity scores on local pose-based BBoxes contribute to these results. Since the general-purpose person ReId problem is significantly harder and requires more sophisticated computational pipelines~\cite{zhao2017deeply,li2014deepreid}, our proposed module seems to take advantage of the body-pose consistency across viewpoints for a faster run-time.

\vspace{1mm}
\textbf{Number of humans and relative viewing angle}: We observed a couple of other practical issues during the field experiments. First, the presence of multiple humans in the scene helps to ensure reliable pose estimation performance. We found that two or more mutually visible humans are ideal for establishing a large pool of reliable correspondences. Additionally, the pose estimation performance is affected by the relative viewing angle; specifically, it often fails to find correct associations when the $\angle$\textit{leader-human-follower} is larger than (approximately) $135^{\circ}$. This results in a situation where the robots are exclusively looking at opposite sides of the person without enough common key-points. Moreover, other than temporal lags, we did not observe significant deviations in pose estimation performance with an increasing number of robots within this viewing angle; note that we used up to three follower robots in our experiments.

\section{Conclusions and Future Work}
In this paper, we explore the feasibility of using human body-poses as markers to establish reliable multi-view geometric correspondences and to eventually solve the robot-to-robot relative pose estimation problem. First, we use OpenPose for extracting the pose-based 2D key-points pertaining to the humans in the scene. Then we associate the humans seen from multiple views using an efficient person re-identification model. Subsequently, we refine the key-point correspondences using an iterative optimization algorithm based on their local structural similarities in the image space. Finally, we use the 3D locations of the key-points (triangulated by the leader robot) and their corresponding 2D projections (seen by the follower robot) to formulate a PnP problem and solve for the unknown pose of the follower robot relative to the leader. We perform extensive experiments in terrestrial and underwater environments to investigate the applicability of the proposed relative pose estimation method; the experimental results validate its effectiveness both for 2D and 3D robots. We also discuss the relevant operational challenges and propose efficient solutions to deal with them. In the future, we seek to improve the end-to-end run-time of the proposed system and plan to use it in practical applications such as multi-robot convoying and cooperative source-to-destination planning. 

\section{Acknowledgments}
We would like to thank Hyun Soo Park (Assistant Professor, University of Minnesota) for his valuable insights which immensely enriched this paper. We are grateful to the Bellairs Research Institute of Barbados for providing us with the facilities for field experiments; we also acknowledge our colleagues at the IRVLab and the participants of the $2019$ Marine Robotics Sea Trials for their assistance in collecting data and conducting the experiments.

\addtolength{\textheight}{-0cm}


\bibliographystyle{IEEEtran}
\bibliography{refs}

\begin{thebibliography}{10}
\providecommand{\url}[1]{#1}
\csname url@samestyle\endcsname
\providecommand{\newblock}{\relax}
\providecommand{\bibinfo}[2]{#2}
\providecommand{\BIBentrySTDinterwordspacing}{\spaceskip=0pt\relax}
\providecommand{\BIBentryALTinterwordstretchfactor}{4}
\providecommand{\BIBentryALTinterwordspacing}{\spaceskip=\fontdimen2\font plus
\BIBentryALTinterwordstretchfactor\fontdimen3\font minus
  \fontdimen4\font\relax}
\providecommand{\BIBforeignlanguage}[2]{{%
\expandafter\ifx\csname l@#1\endcsname\relax
\typeout{** WARNING: IEEEtran.bst: No hyphenation pattern has been}%
\typeout{** loaded for the language `#1'. Using the pattern for}%
\typeout{** the default language instead.}%
\else
\language=\csname l@#1\endcsname
\fi
#2}}
\providecommand{\BIBdecl}{\relax}
\BIBdecl

\bibitem{rekleitis2002multi}
I.~M. Rekleitis, G.~Dudek, and E.~E. Milios, ``{Multi-robot Cooperative
  Localization: A Study of Trade-offs Between Efficiency and Accuracy},'' in
  \emph{{International Conference on Intelligent Robots and Systems (IROS)}},
  vol.~3.\hskip 1em plus 0.5em minus 0.4em\relax IEEE/RSJ, 2002, pp.
  2690--2695.

\bibitem{se2005vision}
S.~Se, D.~G. Lowe, and J.~J. Little, ``{Vision-based Global Localization and
  Mapping for Mobile Robots},'' \emph{{Transactions on Robotics (TRO)}},
  vol.~21, no.~3, pp. 364--375, 2005.

\bibitem{zhou2008robot}
X.~S. Zhou and S.~I. Roumeliotis, ``{Robot-to-robot Relative Pose Estimation
  from Range Measurements},'' \emph{{Transactions on Robotics (TRO)}}, vol.~24,
  no.~6, pp. 1379--1393, 2008.

\bibitem{valgren2010sift}
C.~Valgren and A.~J. Lilienthal, ``{SIFT, SURF \& Seasons: Appearance-based
  Long-term Localization in Outdoor Environments},'' \emph{{Robotics and
  Autonomous Systems}}, vol.~58, no.~2, pp. 149--156, 2010.

\bibitem{damron2018underwater}
H.~Damron, A.~Q. Li, and I.~Rekleitis, ``{Underwater Surveying via Bearing only
  Cooperative Localization},'' in \emph{{International Conference on
  Intelligent Robots and Systems (IROS)}}.\hskip 1em plus 0.5em minus
  0.4em\relax IEEE/RSJ, 2018, pp. 3957--3963.

\bibitem{sattar2008enabling}
J.~Sattar, G.~Dudek, O.~Chiu, I.~Rekleitis, P.~Giguere, A.~Mills, N.~Plamondon,
  C.~Prahacs, Y.~Girdhar, M.~Nahon \emph{et~al.}, ``{Enabling Autonomous
  Capabilities in Underwater Robotics},'' in \emph{{International Conference on
  Intelligent Robots and Systems (IROS)}}.\hskip 1em plus 0.5em minus
  0.4em\relax IEEE/RSJ, 2008, pp. 3628--3634.

\bibitem{islam2018understanding}
M.~J. Islam, M.~Ho, and J.~Sattar, ``{Understanding Human Motion and Gestures
  for Underwater Human-Robot Collaboration},'' \emph{{Journal of Field Robotics
  (JFR)}}, pp. 1--23, 2018.

\bibitem{islam2018person}
M.~J. Islam, J.~Hong, and J.~Sattar, ``{Person-following by Autonomous Robots:
  A Categorical Overview},'' \emph{International Journal of Robotics Research
  (IJRR)}, vol.~38, no.~14, pp. 1581--1618, 2019.

\bibitem{kummerle2013navigation}
R.~K{\"u}mmerle, M.~Ruhnke, B.~Steder, C.~Stachniss, and W.~Burgard, ``{A
  Navigation System for Robots Operating in Crowded Urban Environments},'' in
  \emph{{International Conference on Robotics and Automation (ICRA)}}.\hskip
  1em plus 0.5em minus 0.4em\relax IEEE, 2013, pp. 3225--3232.

\bibitem{zheng2013revisiting}
Y.~Zheng, Y.~Kuang, S.~Sugimoto, K.~Astrom, and M.~Okutomi, ``{Revisiting the
  PnP Problem: A Fast, General and Optimal Solution},'' in \emph{{International
  Conference on Computer Vision (ICCV)}}.\hskip 1em plus 0.5em minus
  0.4em\relax IEEE, 2013, pp. 2344--2351.

\bibitem{cao2017realtime}
Z.~Cao, T.~Simon, S.-E. Wei, and Y.~Sheikh, ``{Realtime Multi-person 2d Pose
  Estimation using Part Affinity Fields},'' in \emph{{Conference on Computer
  Vision and Pattern Recognition (CVPR)}}.\hskip 1em plus 0.5em minus
  0.4em\relax IEEE, 2017, pp. 7291--7299.

\bibitem{trawny2010global}
N.~Trawny and S.~I. Roumeliotis, ``{On the Global Optimum of Planar,
  Range-based Robot-to-robot Relative Pose Estimation},'' in
  \emph{{International Conference on Robotics and Automation (ICRA)}}.\hskip
  1em plus 0.5em minus 0.4em\relax IEEE, 2010, pp. 3200--3206.

\bibitem{zhou2011determining}
X.~S. Zhou and S.~I. Roumeliotis, ``{Determining the Robot-to-robot 3D Relative
  Pose using Combinations of Range and Bearing Measurements (Part II)},'' in
  \emph{{International Conference on Robotics and Automation (ICRA)}}.\hskip
  1em plus 0.5em minus 0.4em\relax IEEE, 2011, pp. 4736--4743.

\bibitem{trawny2010interrobot}
N.~Trawny, X.~S. Zhou, K.~Zhou, and S.~I. Roumeliotis, ``{Inter-robot
  transformations in 3D},'' \emph{{Transactions on Robotics (TRO)}}, vol.~26,
  no.~2, pp. 226--243, 2010.

\bibitem{wang19923d}
J.~Wang and W.~J. Wilson, ``{3D Relative Position and Orientation Estimation
  using Kalman Filter for Robot Control},'' in \emph{{International Conference
  on Robotics and Automation (ICRA)}}.\hskip 1em plus 0.5em minus 0.4em\relax
  IEEE, 1992, pp. 2638--2645.

\bibitem{janabi2010kalman}
F.~Janabi-Sharifi and M.~Marey, ``{A Kalman-filter-based Method for Pose
  estimation in Visual Servoing},'' \emph{{Transactions on Robotics (TRO)}},
  vol.~26, no.~5, pp. 939--947, 2010.

\bibitem{fischler1981random}
M.~A. Fischler and R.~C. Bolles, ``{Random Sample Consensus: A Paradigm for
  Model Fitting with Applications to Image Analysis and Automated
  Cartography},'' \emph{{Communications of the ACM}}, vol.~24, no.~6, pp.
  381--395, 1981.

\bibitem{rekleitis2006simultaneous}
I.~Rekleitis, D.~Meger, and G.~Dudek, ``{Simultaneous Planning, Localization,
  and Mapping in a Camera Sensor Network},'' \emph{{Robotics and Autonomous
  Systems}}, vol.~54, no.~11, pp. 921--932, 2006.

\bibitem{gkioxari2014using}
G.~Gkioxari, B.~Hariharan, R.~Girshick, and J.~Malik, ``{Using K-poselets for
  Detecting People and Localizing their Keypoints},'' in \emph{{Conference on
  Computer Vision and Pattern Recognition (CVPR)}}.\hskip 1em plus 0.5em minus
  0.4em\relax IEEE, 2014, pp. 3582--3589.

\bibitem{pishchulin2012articulated}
L.~Pishchulin, A.~Jain, M.~Andriluka, T.~Thorm{\"a}hlen, and B.~Schiele,
  ``{Articulated People Detection and Pose Estimation: Reshaping the Future},''
  in \emph{{Conference on Computer Vision and Pattern Recognition
  (CVPR)}}.\hskip 1em plus 0.5em minus 0.4em\relax IEEE, 2012, pp. 3178--3185.

\bibitem{pishchulin2016deepcut}
L.~Pishchulin, E.~Insafutdinov, S.~Tang, B.~Andres, M.~Andriluka, P.~V. Gehler,
  and B.~Schiele, ``{DeepCut: Joint Subset Partition and Labeling for Multi
  Person Pose Estimation},'' in \emph{{Conference on Computer Vision and
  Pattern Recognition (CVPR)}}.\hskip 1em plus 0.5em minus 0.4em\relax IEEE,
  2016, pp. 4929--4937.

\bibitem{ferrari2008progressive}
V.~Ferrari, M.~Marin-Jimenez, and A.~Zisserman, ``{Progressive Search Space
  Reduction for Human Pose Estimation},'' in \emph{{Conference on Computer
  Vision and Pattern Recognition (CVPR)}}.\hskip 1em plus 0.5em minus
  0.4em\relax IEEE, 2008, pp. 1--8.

\bibitem{andriluka2009pictorial}
M.~Andriluka, S.~Roth, and B.~Schiele, ``{Pictorial Structures Revisited:
  People Detection and Articulated Pose Estimation},'' in \emph{{Conference on
  Computer Vision and Pattern Recognition (CVPR)}}.\hskip 1em plus 0.5em minus
  0.4em\relax IEEE, 2009, pp. 1014--1021.

\bibitem{johnson2011learning}
S.~Johnson and M.~Everingham, ``{Learning Effective Human Pose Estimation from
  Inaccurate Annotation},'' in \emph{{Conference on Computer Vision and Pattern
  Recognition (CVPR)}}.\hskip 1em plus 0.5em minus 0.4em\relax IEEE, 2011, pp.
  1465--1472.

\bibitem{pishchulin2013poselet}
L.~Pishchulin, M.~Andriluka, P.~Gehler, and B.~Schiele, ``{Poselet Conditioned
  Pictorial Structures},'' in \emph{{Conference on Computer Vision and Pattern
  Recognition (CVPR)}}.\hskip 1em plus 0.5em minus 0.4em\relax IEEE, 2013, pp.
  588--595.

\bibitem{toshev2014deeppose}
A.~Toshev and C.~Szegedy, ``{DeepPose: Human Pose Estimation via Deep Neural
  Networks},'' in \emph{{Conference on Computer Vision and Pattern Recognition
  (CVPR)}}.\hskip 1em plus 0.5em minus 0.4em\relax IEEE, 2014, pp. 1653--1660.

\bibitem{ramakrishna2014pose}
V.~Ramakrishna, D.~Munoz, M.~Hebert, J.~A. Bagnell, and Y.~Sheikh, ``{Pose
  Machines: Articulated Pose Estimation via Inference Machines},'' in
  \emph{{European Conference on Computer Vision (ECCV)}}.\hskip 1em plus 0.5em
  minus 0.4em\relax Springer, 2014, pp. 33--47.

\bibitem{mead2017autonomous}
R.~Mead and M.~J. Matari{\'c}, ``{Autonomous Human--robot Proxemics: Socially
  aware Navigation based on Interaction Potential},'' \emph{{Autonomous
  Robots}}, vol.~41, no.~5, pp. 1189--1201, 2017.

\bibitem{montemerlo2002conditional}
M.~Montemerlo, S.~Thrun, and W.~Whittaker, ``{Conditional Particle Filters for
  Simultaneous Mobile Robot Localization and People-tracking},'' in
  \emph{{International Conference on Robotics and Automation (ICRA)}},
  vol.~1.\hskip 1em plus 0.5em minus 0.4em\relax IEEE, 2002, pp. 695--701.

\bibitem{mainprice2013human}
J.~Mainprice and D.~Berenson, ``{Human-robot Collaborative Manipulation
  Planning using Early Prediction of Human Motion},'' in \emph{{International
  Conference on Intelligent Robots and Systems (IROS)}}.\hskip 1em plus 0.5em
  minus 0.4em\relax IEEE/RSJ, 2013, pp. 299--306.

\bibitem{lei2015whole}
J.~Lei, M.~Song, Z.-N. Li, and C.~Chen, ``{Whole-body Humanoid Robot Imitation
  with Pose Similarity Evaluation},'' \emph{{Signal Processing}}, vol. 108, pp.
  136--146, 2015.

\bibitem{Jetson}
{NVIDIA\texttrademark}, ``{Embedded Computing Boards},''
  \url{developer.nvidia.com/embedded/jetson-tx2}, 2014, accessed: 8-2-2019.

\bibitem{ahmed2015improved}
E.~Ahmed, M.~Jones, and T.~K. Marks, ``{An Improved Deep Learning Architecture
  for Person Re-identification},'' in \emph{{Conference on Computer Vision and
  Pattern Recognition (CVPR)}}.\hskip 1em plus 0.5em minus 0.4em\relax IEEE,
  2015, pp. 3908--3916.

\bibitem{li2014deepreid}
W.~Li, R.~Zhao, T.~Xiao, and X.~Wang, ``{Deepreid: Deep Filter Pairing Neural
  Network for Person Re-identification},'' in \emph{{Conference on Computer
  Vision and Pattern Recognition (CVPR)}}.\hskip 1em plus 0.5em minus
  0.4em\relax IEEE, 2014, pp. 152--159.

\bibitem{wang2004image}
Z.~Wang, A.~C. Bovik, H.~R. Sheikh, E.~P. Simoncelli \emph{et~al.}, ``{Image
  Quality Assessment: from Error Visibility to Structural Similarity},''
  \emph{{Transactions on Image Processing (TIP)}}, vol.~13, no.~4, pp.
  600--612, 2004.

\bibitem{avanaki2009exact}
A.~N. Avanaki, ``{Exact Global Histogram Specification Optimized for Structural
  Similarity},'' \emph{{Optical Review}}, vol.~16, no.~6, pp. 613--621, 2009.

\bibitem{otero2014solving}
D.~Otero and E.~R. Vrscay, ``{Solving Optimization Problems that Employ
  Structural Similarity as the Fidelity Measure},'' in \emph{{International
  Conference on Image Processing, Computer Vision, and Pattern Recognition
  (IPCV)}}, 2014, p.~1.

\bibitem{hartley2003multiple}
R.~Hartley and A.~Zisserman, \emph{{Multiple View Geometry in Computer
  Vision}}.\hskip 1em plus 0.5em minus 0.4em\relax {Cambridge University
  Press}, 2003.

\bibitem{zhao2017deeply}
L.~Zhao, X.~Li, Y.~Zhuang, and J.~Wang, ``{Deeply-learned Part-aligned
  Representations for Person Re-identification},'' in \emph{{IEEE International
  Conference on Computer Vision (ICCV)}}, 2017, pp. 3219--3228.

\bibitem{zheng2011person}
W.-S. Zheng, S.~Gong, and T.~Xiang, ``{Person Re-identification by
  Probabilistic Relative Distance Comparison},'' in \emph{{Conference on
  Computer Vision and Pattern Recognition (CVPR)}}.\hskip 1em plus 0.5em minus
  0.4em\relax IEEE, 2011, pp. 649--656.

\bibitem{FLANN}
{OpenCV}, ``{Fast Library for Approximate Nearest Neighbors (FLANN)-based 2D
  Feature Matching Algorithm},''
  \url{https://docs.opencv.org/3.4/d5/d6f/tutorial_feature_flann_matcher.html},
  2018, accessed: 6-20-2020.

\bibitem{shkurti2017underwater}
F.~Shkurti, W.-D. Chang, P.~Henderson, M.~J. Islam, J.~C.~G. Higuera, J.~Li,
  T.~Manderson, A.~Xu, G.~Dudek, and J.~Sattar, ``{Underwater Multi-Robot
  Convoying using Visual Tracking by Detection},'' in \emph{{International
  Conference on Intelligent Robots and Systems (IROS)}}.\hskip 1em plus 0.5em
  minus 0.4em\relax IEEE/RSJ, 2017.

\bibitem{RosTime}
{ROS.org}, ``{ROS Time Synchronizer},''
  \url{http://wiki.ros.org/message_filters}, 2018, accessed: 6-20-2020.

\end{thebibliography}

\end{document}